\documentclass[journal]{IEEEtran}

\usepackage{xcolor}
\usepackage{graphicx}
\usepackage{amsmath}
\usepackage{booktabs}
\usepackage{multirow}
\usepackage{color, colortbl}
\usepackage{graphicx,dblfloatfix}
\usepackage[noadjust]{cite}

\begin{document}

\title{Learning Woody Clearing with Loss Alignment for Zero-Shot Regrowth and Woody Segmentation}

\author{Kal~Backman,
        Jared~Wood and
        Adam~Roff
\thanks{The authors are with the New South Wales Department of Climate Change, Energy, the Environment and Water, Parramatta, 2150, NSW, Australia}
\thanks{K. Backman is also affiliated with Monash University, Clayton, 3800, Vic, Australia}
\thanks{A. Roff is also affiliated with Earth Observation Lab, University of New England, Armidale, 2351, NSW, Australia and the Conservation Science Research Group, University of Newcastle, Callaghan, 2308, NSW, Australia}}

\markboth{Journal of \LaTeX\ Class Files,~Vol.~X, No.~X, Month~20XX}%
{Shell \MakeLowercase{\textit{et al.}}: Bare Demo of IEEEtran.cls for Journals}
%



\maketitle

\begin{abstract}
Detecting woody clearing is vital for managing biodiversity. Deep learning models can detect change in woody vegetation from bitemporal remote sensing imagery, however generated products may not meet end-user specifications due to unaligned loss definitions. Further limitations of deep learning models are the reliance on large datasets which can be difficult to attain for spatially rare and ambiguous events such as regrowth detection.
In this work we train a model to detect woody change using bitemporal Sentinel-2 imagery consisting of 7 years’ worth of annual imagery across the state of New South Wales, Australia. To align the objective of the model with end-user metrics, we introduce the loss scaling coefficient \(\alpha\) which transforms the objective to optimize for specific \(F_{\beta}\) scores. Introducing \(\alpha\) was found to increase precision by 1.85\(\times\) or recall by 1.12\(\times\).
We propose input imagery augmentation and generation techniques that allow the woody change detection model to zero-shot transfer to regrowth and woody segmentation tasks.
For woody segmentation, image generation techniques using activation maximization with low \(\alpha\) values for stability and image generation techniques derived from handcrafted features utilizing a mosaic of clearing patches and artificial trees for contextual grounding were found to outperform prior woody segmentation works of the study area, reducing the overall error by up to 18.2\%.
For zero-shot woody regrowth, creating pseudo-post and prior images resulted in the model achieving an F1 score of 0.845, creating a foundation for future regrowth detection work.
\end{abstract}


\IEEEpeerreviewmaketitle

\section{Introduction}
Land clearing is the leading cause of terrestrial biodiversity loss \cite{diaz2019pervasive, legge2023loss, jaureguiberry2022direct, davison2021land}. 
With the continual growth in the demand for agricultural land \cite{williams2021proactive, tilman2017future}, natural resources \cite{diaz2019pervasive} and urbanization \cite{kati2021biodiversity, de2020vulnerability}, it is more imperative than ever that we enact land clearing reform to protect biodiversity \cite{fletcher2024earth, reside2017ecological}.
In order to protect current biodiversity, timely and accurate data of land clearing trends and extent are required to make informed decisions \cite{mccord2021provoking, farley2018situating}.

Land clearing detection within the context of remote sensing stems from the broader domain of change detection, which aims to denote changes in the semantic classes of pixels across time within remote sensing imagery. Change detection has shown success in many applications including post disaster assessment \cite{brunner2010earthquake, zheng2021building, qing2022operational}, urbanization and infrastructure expansion \cite{papadomanolaki2021deep, guo2021deep, huang2020automatic} and land cover / land use change monitoring \cite{chen2024objformer, seyam2023identifying, zhu2022land, das2022land}. 

Traditional change detection methods tend to operate on a single pixel level, comparing spectral indices and individual bands across the time-series to make decisions \cite{scarth2008assimilation, coppin1996digital, gianinetto2011mapping, xian2010updating}. However such traditional single pixel methods ignore key contextual information, resulting in noisy predictions that contain holes in continuous regions denoted as change \cite{radke2005image, bontemps2008object}. 

Deep learning approaches circumvent this singular pixel context window limitation by pooling information spatially at different spatial resolutions to attain a greater context window. Common deep learning methods focus on using convolutional neural networks (CNN) \cite{isaienkov2020deep, lin2022transition, lv2022simple}, self-attention \cite{li2022densely, fang2023changer, chen2021remote} or Mamba \cite{zhang2025cdmamba, paranjape2025mamba} architectures to detect change within an image. 

The behavior of deep learning approaches are defined by the dataset they are trained on and the loss function used to map input imagery to output segmentation masks. The cross-entropy loss is a commonly used loss function for remote sensing change detection tasks \cite{fang2023changer, chen2021remote, pan2024m, mou2018learning}. However for change detection tasks focusing on heavily class imbalanced datasets, a combination of the cross-entropy loss and dice loss is preferred \cite{isaienkov2020deep, li2022densely, li2021combined} due to the dice loss not being strongly biased towards the majority class \cite{milletari2016v}.

However such loss functions may not produce input-output mappings that are aligned with the objectives of the end-user, where an essential quality for successful remote sensing products is that they provide the relevant and accurate insights to the user \cite{kennedy2009remote, czaplewski2003classification}. 
These relevant end-user accuracy definitions can be biased towards optimizing a specific quality such as recall at the cost of precision, as seen in operational programs utilizing region proposal models for spatially rare events that require skilled interpreters to edit, verify and evaluate all published remote sensing products \cite{qdec2018slats}.
By default, the aforementioned loss functions are unable to bias the model to prioritize recall or precision over the other. Although confidence thresholding can be used as a post processing operation to bias predictions, neural networks are known to suffer from overconfidence issues \cite{hwang2024overcoming, aliferis2024overfitting, wei2022mitigating}. This overconfidence causes the bulk of predictions to be concentrated in the tails of the confidence distribution, leading to confidence thresholds with extremely low margins of error to attain the desired model performance. 

Another issue with deep learning approaches for change detection is the vast amounts of data required to train a suitable model. This is further challenging for change detection tasks that aim to detect rare events due to the limited data available. Although transfer learning can be used to minimize data requirements \cite{ma2024transfer, weiss2016survey}, an initial dataset is still required.
Obtaining an initial dataset for tasks such as woody regrowth is challenging, unlike woody clearing which often involves discrete abrupt events, woody regrowth is a visually subtle and continuous process occurring over several years. 
Despite decades of efforts, comprehensive and consistent regrowth mapping remains limited \cite{souza2020reconstructing, qdec2024slats}, preventing deep learning approaches from being applied to regrowth mapping due to the absence of sufficient training data \cite{yuan2021review}. 

Zero-shot learning aims to circumvent the initial dataset requirement by transferring trained models to unseen tasks without additional training or data. Zero-shot learning allows deep learning models to complete tasks where training data is insufficient such as woody regrowth detection by leveraging the abundance of similarly distributed datasets such as woody clearing.

In this work we aim to train a deep learning model that detects woody vegetation clearing using bitemporal geospatially aligned Sentinel-2 imagery. For the purpose of this work, woody vegetation is defined as vegetation over the height of 2m. The objective of the model is to optimize for \(F_\beta\) scores of large \(\beta\) values, i.e. prioritize recall of woody clearing over precision. To align the model’s learning objective with the end user’s objective of maximizing recall, the loss scaling coefficient \(\alpha\) was introduced to allow the learning process to explicitly target a specific metric during training.

The proposed model comprises of a CNN twin encoder decoder network that outputs a segmentation mask denoting woody vegetation that is absent in the post image relative to the prior image.
To further improve woody vegetation monitoring, the proposed model was transferred in a zero-shot manner to woody regrowth and woody segmentation tasks by generating and augmenting input imagery to extract latent information from within the network.
The proposed work aims to build a foundation for further regrowth detection work by establishing a method that can initially propose regrowth regions to address the difficulty in attaining woody regrowth data.    

The main contributions of the proposed work are:
\begin{enumerate}
\item The development of a woody clearing network trained and validated on a large dataset consisting of 7 years’ worth of imagery covering a landmass of 801,137 km\(^2\) and containing 54 billion data points.
\item The proposal of the loss scaling coefficient \(\alpha\) that modifies the binary cross entropy and dice loss to align the training objective towards end user orientated metrics. Capable of increasing precision or recall by a factor of 1.85\(\times\) or 1.12\(\times\) respectively in the woody clearing task and increasing the F1 score by 1.32\(\times\) for the zero-shot woody segmentation task.
\item Input imagery augmentation and generation techniques that allow a model trained purely on woody change detection data to zero-shot transfer to woody regrowth and woody segmentation tasks. Zero-shot woody segmentation was found to outperform prior woody segmentation maps of the study area with a reduction of the overall error by 18.2\%, and the first work to build the foundation for woody regrowth for the study area due to the difficulties of establishing a curated dataset.        
\end{enumerate}

\section{Methodology}
This section introduces the woody change detection architecture, the proposed loss function and the woody change detection dataset used for training and validation.

\subsection{Model architecture}
\begin{figure}[b]
\centering
\includegraphics[width=1.0\columnwidth]{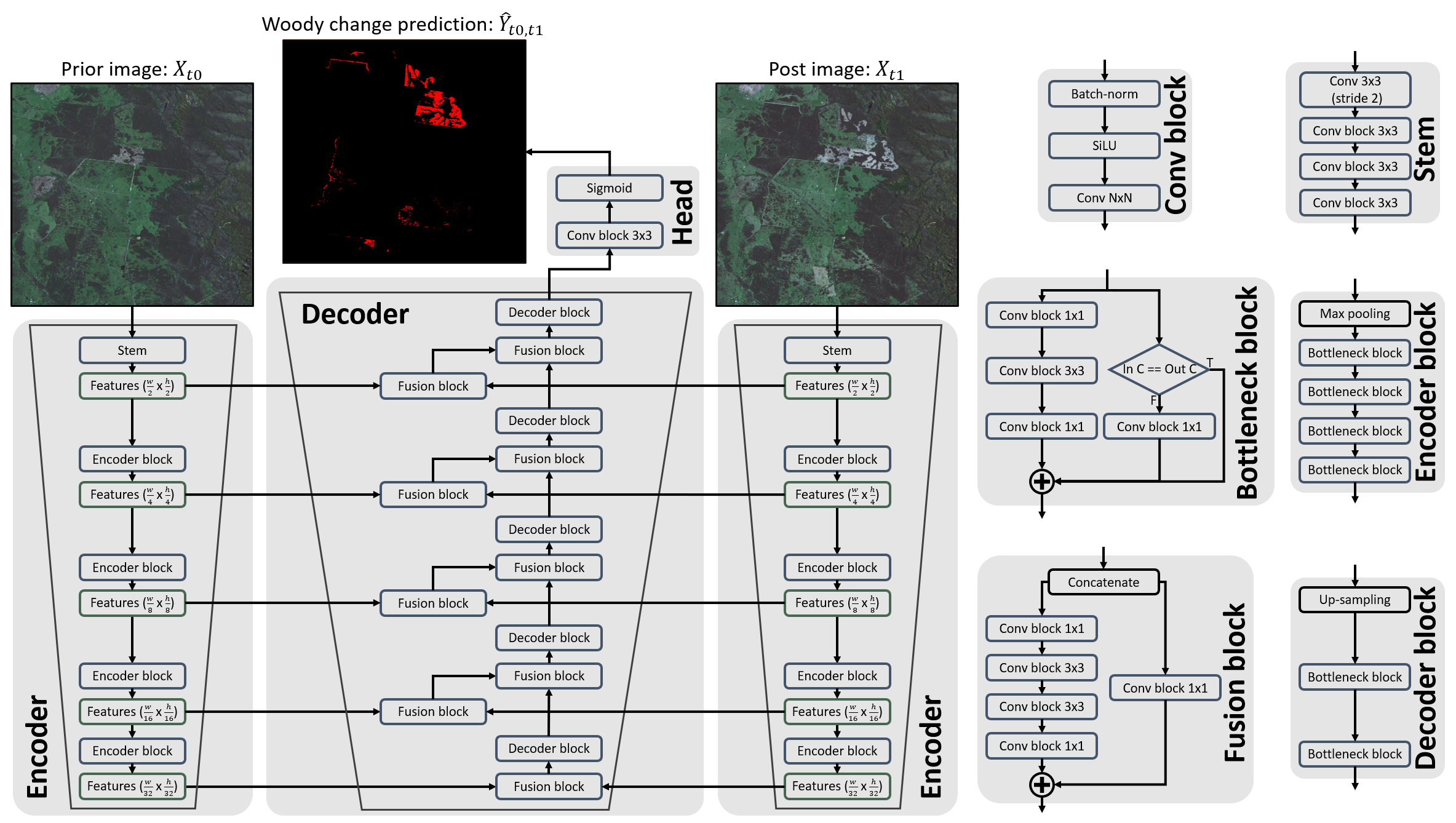}
\caption{Overview of the woody change detection model architecture. The encoder individually extracts features from the prior and post images at different resolution scales using convolution-based bottleneck blocks. The extracted features are subsequently fused to encode woody change at each resolution scale. The lowest resolution feature is progressively upscaled and fused with the subsequent higher resolution features to produce woody change detection predictions at an identical resolution as the input imagery.}
\label{modelOverview}
\end{figure}

The woody change detection architecture follows an encoder-decoder architecture comprising of three components: the encoder, the decoder and the head. 
The objective of the encoder is to extract multiscale features from bitemporal images pairs to describe the structure of the scene. 
The objective of the decoder is to fuse information across different temporal and spatial scales to detect regions of woody change.
The objective of the head is to produce output confidence scores that denote whether a corresponding pixel in the bitemporal input pair contains woody clearing.
An overview of the model architecture can be seen in Fig.~\ref{modelOverview}.

\subsubsection{Encoder}
The encoder employs a Siamese convolutional neural network that independently extracts multiscale features from the bitemporal image pair. The encoder receives two spatially aligned images: \(X_{t0}\) \& \(X_{t1}\), where \(X_{t0}\) denotes the image taken prior to \(X_{t1}\) while \(X_{t1}\) denotes the image post of \(X_{t0}\).
Both \(X_{t0}\) \& \(X_{t1}\) are passed through identically weighted backbones which output a total of 5 feature maps denoted as \(F_{ti}^{j}\), where \(i\) refers to the input imagery time i.e. \(X_{ti}\) and \(j\) denoting the depth of the feature map. Features [\(F_{ti}^{1}\), \(F_{ti}^{2}\), \(F_{ti}^{3}\), \(F_{ti}^{4}\), \(F_{ti}^{5}\)] contain [64, 128, 256, 512, 1028] output channels at resolution [\(\frac{1}{2}\), \(\frac{1}{4}\), \(\frac{1}{8}\), \(\frac{1}{16}\), \(\frac{1}{32}\)] relative to the input imagery.
 
The initial feature map is produced by the backbone’s stem which receives the input imagery and applies an initial convolutional layer with stride length of 2, followed by 3 consecutive convolutional blocks which comprise of batch-norm, SiLU activation and convolutional layers. Successive feature maps are generated by receiving the prior feature map as input and applying a max pooling operation followed by 4 consecutive bottleneck residual blocks \cite{he2016deep} for parameter efficiency.         

\subsubsection{Decoder}
The decoder receives the 5 feature maps from both bitemporal images (\(F_{t0}^{j}\) \& \(F_{t1}^{j}\)) and initially fuses \(F_{t0}^{5}\) \& \(F_{t1}^{5}\) via a fusion block comprising of an initial concatenation layer followed by a bottleneck residual block with a convolutional block to facilitate a skip connection due to the reduction in channel dimensionality. The result is fed through the decoder block which comprises of an initial up-sampling operation to match the resolution of the next feature map, followed by 2 consecutive bottleneck residual blocks, where the intermediate output is denoted as \(\bar{F}_{t0,t1}^{5}\).
The decoder subsequently fuses the next set of bitemporal features which are then fused with the prior feature stage i.e. Fuse(Fuse(\(F_{t0}^{j-1}\), \(F_{t1}^{j-1}\)), \(\bar{F}_{t0,t1}^{j}\)). The fused features are passed through the decoder block and the process is repeated until the subsequent feature map resolution matches the original input image resolution.

\subsubsection{Head}
The head receives the final output from the decoder and applies a single convolutional block followed by a sigmoid activation to generate confidence scores that a pixel in image \(X_{t1}\) contains woody clearing relative to image \(X_{t0}\) which is denoted as \(\hat{Y}_{t0,t1}\). Output \(\hat{Y}_{t0,t1}\) can be subsequently thresholded to a set confidence value to produce a binary mask indicating woody clearing.

\subsection{Loss function}
The greatest challenge in training a network to segment woody clearing from remote sensing imagery is the extreme class imbalance from the scarcity of woody clearing. Woody clearing can constitute 0.05\% of total pixels as seen in section \ref{section_traingDataset}, resulting in a class imbalance of 1 positive pixel per 2,000 negative pixels. 
Despite its popularity for segmentation tasks \cite{fang2023changer, chen2021remote, pan2024m, mou2018learning}, the binary cross-entropy loss heavily biases the objective function towards the majority class due to equally weighting the minimization of log-probabilities across all pixels. This leads to the network getting stuck in a sub-optimal local minima due to prioritizing the minimization of a large sum of small errors in the majority class, over a small sum of large errors in the minority class. 
The dice loss on the other hand computes the overlap between the target and the prediction, making it more robust to class imbalances \cite{milletari2016v} due to penalizing the network on false positives and false negatives. However the dice loss is prone to over fixating on erroneous labels in batches containing few positive samples.
Therefore a combination of the binary cross-entropy loss (\(L_{BCE}\)) and dice loss (\(L_{Dice}\)) is commonly used for class imbalanced applications \cite{isaienkov2020deep, li2022densely, li2021combined}, which can further be weighted for cases of extreme class disparity:
\begin{equation}
{L_{Total} = w_{BCE} \cdot L_{BCE} + w_{Dice} \cdot L_{Dice}}.
\end{equation}

The dice loss is defined as:
\begin{equation}
L_{Dice} = 1 - \frac{2 \sum\limits_{i=1}^{n} {Y}_{i} \hat{Y}_i}{\sum\limits_{i=1}^{n} Y_i + \sum\limits_{i=1}^{n} \hat{Y}_i},
\end{equation}
which can be further generalized to the expression:
\begin{equation}\label{Equation_weightedDice}
L_{Dice} = 1 - \frac{2 \sum\limits_{i=1}^{n} Y_{i} \hat{Y}_i w_i + \epsilon}{\sum\limits_{i=1}^{n} Y_i w_i + \sum\limits_{i=1}^{n} \hat{Y}_i w_i + \epsilon},
\end{equation}
where \(w_i \in\![0, 1]\) denotes a weighting score which can be used to reflect the confidence in the validity of the given label \(Y_i\) while \(\epsilon = 0.001\) is used for stability to prevent small division errors.
Similarly the generalized binary cross-entropy loss can be expressed as:
\begin{equation}\label{Equation_weightedBCE}
L_{BCE} = -\frac{1}{n}\sum\limits_{i=1}^{n}( Y_i \log(\hat{Y}_i) + (1 - Y_i) \log(1-\hat{Y}_i)) w_i.  
\end{equation}  

Observing equations \ref{Equation_weightedDice} \& \ref{Equation_weightedBCE}, it can be seen that both loss functions equally weight the optimization of both precision and recall metrics.
For the dice loss this is achieved via dividing by the sum of ground truths (\(Y_i\)) which restricts the prevalence of false negatives, therefore boosting recall. Dividing by the sum of predictions (\(\hat{Y}_i\)) restricts the prevalence of false positives, thereby boosting precision.
For the binary cross-entropy loss this is achieved via multiplying the ground truth with the log probability of the prediction (\(Y_i \log(\hat{Y}_i)\)) which aims to enhance true positives, therefore boosting recall. The subsequent term (\((1 - Y_i) \log(1-\hat{Y}_i)\)) aims to enhance true negatives, therefore boosting precision.

The equal optimization of both precision and recall aims to maximize the \(F_1\) score, a common metric to rank the performance of segmentation models and is defined in equation~\ref{Equation_F1} in section \ref{section_clearingDetectionEval}.
However for certain applications it may be crucial to prioritize precision or recall over the other. In such cases, higher confidence thresholds can be applied to attain greater precision whilst the inverse can be used to enhance recall. 
However neural networks are known to suffer from overconfidence issues \cite{hwang2024overcoming, aliferis2024overfitting, wei2022mitigating}, leading to an invalid representation of the true confidence and therefore making threshold adjustments unreliable.
Instead, defining a loss function that allows users to explicitly bias the learning objective towards precision or recall to suit their application is required.

To allow for a tunable loss function such as to optimize for specific \(F_{\beta}\) scores, we introduce the tunable parameter \(\alpha \in\!(0, \infty)\) into the loss functions depicted in equations  \ref{Equation_weightedDice} \& \ref{Equation_weightedBCE}:
\begin{equation}\label{Equation_alphaDice}
L_{Dice} = 1 - \frac{2 \sum\limits_{i=1}^{n} Y_{i} \hat{Y}_i w_i + \epsilon}{\sum\limits_{i=1}^{n} Y_i w_i \alpha + \sum\limits_{i=1}^{n} \hat{Y}_i w_i \frac{1}{\alpha} + \epsilon},
\end{equation}
\begin{equation}\label{Equation_alphaBCE}
L_{BCE} = -\frac{1}{n}\sum\limits_{i=1}^{n}( Y_i \log(\hat{Y}_i) \alpha + (1 - Y_i) \log(1-\hat{Y}_i) \frac{1}{\alpha}) w_i.  
\end{equation}  
Terms that favour recall are multiplied by \(\alpha\) whilst terms that favour precision are multiplied by \(\frac{1}{\alpha}\), such that for values \(\alpha > 1\), a greater priority is given to enhancing recall whilst for values \(\alpha < 1\), precision is favored. The use of \(\alpha\) and \(\frac{1}{\alpha}\) allows for modifying each term by an identical factor to simplify the tuning process. 

\subsection{Data}
\subsubsection{Raw input imagery}
The input imagery used to train and validate the woody clearing architecture was based on 10m Sentinel-2 imagery captured across mainland New South Wales, Australia covering a total landmass of 801,137 km\(^2\) per year. Sentinel-2 imagery was captured across the summer period, where the majority of images were sampled between November to early March across a 7 year period from 2018 to 2024. 

The Sentinel-2 imagery was split into 109 roughly 100km \(\times\) 100km scenes which had radiometric corrections applied to compute the JRSRP Sentinel-2 surface reflectance outlined in \cite{flood2013operational, flood2020comparing}, and cloud masking used to rank image quality. 
For each year, a new post image was selected based on the availability of cloud free imagery, taking the closest image in date to January 1, while the prior image for a given era was selected from the post image of the previous era. Some images were created from a mosaic of two or three images where cloud or incomplete captures meant clean imagery was not available.
The Sentinel-2 imagery was subsequently cropped to a common image size and spatial boundary for each scene, such that all images within a scene could be stacked with each pixel corresponding to an identical physical location across the images. 

\subsubsection{Woody clearing labels}
To generate woody clearing labels, regions of interest were initially proposed via a clearing regression index outlined in \cite{scarth2008assimilation} and \cite{qdec2018slats} which compares individual pixels in the prior and post images to compute a probability that the particular pixel contains woody clearing. The region of interest proposal model was designed to highly emphasize recall at the cost of precision as to not exclude woody clearing events. 

A team of operators manually checked through all proposed regions of interest across the state for a given era and denoted which pixels within these regions of interest contained woody clearing and for what purpose were they cleared for (e.g. agricultural). 
A secondary check was subsequently performed by an independent operator to validate that the indicated woody clearing pixels were correct.
During both manual woody clearing checks, operators are able to reference higher resolution auxiliary data sources to disambiguate woody clearing events.  
The process was repeated yearly for a total of 6 eras' worth of labels beginning from (2018-2019) to (2023-2024) were obtained.

The labels across all eras were subsequently cropped to an identical image and spatial boundary such that they overlap with the clipped Sentinel-2 imagery.

\subsubsection{Training and validation datasets}\label{section_traingDataset}
A dataset was generated to train the woody clearing segmentation model by initially extracting all 10m bands: 2 (blue 490nm), 3 (green 560nm), 4 (red 665nm) and 8 (NIR 842nm) and a single 20m band 12 (SWIR 2190nm) which was resampled to stack on top of the 10m bands.
The extracted bands were normalized between [-1, 1] by computing a histogram and clipping the lowest and highest 2.5\% of values. The new minimum and maximum values were used to linearly scale each band between -1 and 1.
The normalized extracted bands were subsequently split into 400w \(\times\) 400h tiles. The tiles were generated such that for all years for a given scene, each tile could be stacked so that each pixel corresponds to an identical physical location. 

Labels were similarly split into 400w \(\times\) 400h tiles which overlapped with the input imagery. Binary labels were formulated where pixels denoted as clearing were assigned a value of 1 whilst non-clearing pixels were assigned a value of 0. As woody clearing resultant from pine plantations were not required to be reported on, operators did not encode plantation clearing as woody clearing due to time constraints, resulting in woody clearing from plantations erroneously being encoded with a value of 0.

An additional weighting tile was generated to scale the loss functions outlined in equations \ref{Equation_alphaDice} \& \ref{Equation_alphaBCE}. The purpose of the weight tile is to zero weight invalid regions where data may not be present and to weight the loss function to account for the confidence in the label source to minimize the impact of erroneous labels \cite{backman2024bicycle}. As all labels denoted as woody clearing have been manually validated by at least two human operators, it provides a highly trusted point of truth. Regions not flagged by the initial clearing regression index are not required to be manually validated and therefore may contain a greater rate of errors. To account for this, human verified clearing events are assigned a weighting of 1.0, whilst non-verified non clearing events are assigned a weighting of 0.1. To minimize the influence of the incorrect classification of woody clearing deriving from pine plantations, non-clearing pixels which returned a high-probability for clearing from the initial clearing regression index had their weight values further decayed by 50\%. Invalid pixels such as those obstructed by cloud or pixels that fall outside of the boundaries of New South Wales were assigned a value of 0.0 due to being excluded from the labeling process.   

Labels associated with eras (2018-2019) to (2022-2023) were assigned to the training dataset whilst the remaining era (2023-2024) was excluded and reserved for validation purposes only. The total dataset consisted of 284,215 tiles for training and 56,851 for validation, where the woody clearing present in the dataset was 0.05\% of all pixels.

\section{Experiment design}
This section introduces the experiment procedures to validate the proposed woody change detection architecture. 
The initial subsection outlines the model training and evaluation procedure to segment woody clearing.
The subsequent subsection explores the model’s response to different loss scaling coefficients and their ability to optimize for specific \(F_{\beta}\) scores. 
The remaining two subsections explore methods that allow the model to be applied to zero-shot tasks such as woody segmentation and woody regrowth detection without the need for further training.

\subsection{Woody clearing detection}\label{section_clearingDetectionExp}
\subsubsection{Training}\label{Section_clearingDetectionTraining}
The woody change detection model was trained by sampling prior and post image pairs alongside the accompanied label using a batch size of 128. To account for the scarcity of woody clearing within the dataset, 50\% of data points within a batch were randomly sampled from a known set of tiles that contained woody clearing, whilst the remaining 50\% were sampled randomly from the entire training dataset.

The sampled training tiles had a random combination of data augmentation techniques applied. Data augmentation techniques used can be grouped into 3 categories: those that alter the appearance of the base imagery, those that add noise to the imagery and those that geometrically distort the imagery. 
Data augmentation techniques used to alter the appearance of the base imagery include gamma correction transforms to artificially darken or brighten the image and band-shift transforms to artificially bring out different colors within the image. 
Noising transforms include adding normally distributed noise to the image and applying Gaussian blur operations. 
Geometric augmentations include flipping the image along the vertical and horizontal axis, rotating the image, scaling the image to become larger or smaller and shearing the image to skew its perspective.
After data augmentations were applied, a randomly selected point on the image was chosen and cropped to the training input image size of 256w \(\times\) 256h.

The model was trained for a total of 200,000 optimization iterations split into 64 epochs containing 3,125 optimization iterations. Loss scaling coefficient \(\alpha\) was set to 1.0, whilst the binary cross entropy loss scaling coefficient \(w_{BCE}\) was set to 0.1 and the dice loss scaling coefficient \(w_{Dice}\) was set to 1.0. The Adam optimizer was used with an initial learning rate of 2e-4 and was linearly decayed to the final learning rate of 1e-5. Gradient clipping was performed using a maximum gradient norm of 1.0 and gradient scaling for mixed precision layers. The model was trained using a NVIDIA A100 80GB GPU. 

\subsubsection{Evaluation}\label{section_clearingDetectionEval}
The trained model was tasked with generating woody clearing predictions for entire scenes across the (2023-2024) era. The model received an input image size of 1024w \(\times\) 1024h using a batch size of 32. The input image patch was swept across the image using an image overlap of 50\%. A weighted average of prediction confidences was used to fuse overlapping segments, where pixels far away from the center of the image patch were weighted lower than those in the center due to the lack of contextual information around the borders. 

The model was evaluated under 5 metrics: precision, recall, F1-score, F2-score and intersection over union (IoU) which are defined as follows:
\begin{equation}\label{Equation_precision}
\text{Precision} = \frac{TP}{TP + FP}  
\end{equation}  
\begin{equation}\label{Equation_recall}
\text{Recall} = \frac{TP}{TP + FN}  
\end{equation}  
\begin{equation}\label{Equation_F1}
\text{F1} = \frac{2 \cdot \text{Precision} \cdot \text{Recall}}{\text{Precision} + \text{Recall}}  
\end{equation}
\begin{equation}\label{Equation_F2}
\text{F2} = \frac{5 \cdot \text{Precision} \cdot \text{Recall}}{4 \cdot \text{Precision} + \text{Recall}}  
\end{equation}  
\begin{equation}\label{Equation_IoU}
\text{IoU} = \frac{TP}{TP + FP + FN}  
\end{equation}
\(TP\) denotes the number of true positives, where a true positive is defined as a prediction that is above a threshold confidence level that corresponds to a ground truth label of woody change.
\(FP\), \(TN\) and \(FN\) denote the number of false positives, true negatives and false negatives respectively.

To better understand the distribution of errors, the aforementioned metrics were recomputed whilst considering segment matching and distance thresholds. Segment analysis aims to test the model’s ability to detect patches of clearing compared to individual pixels.
Segments were computed by buffering all woody clearing pixels by 20m (2 pixels), where adjacently connected buffered pixels were grouped and assigned unique segment IDs. The segment IDs were subsequently transferred to the unbuffered pixels. Segments are computed for both the model’s predictions and the ground truth labels, resulting in each pixel in the unbuffered prediction and ground truth corresponding to a unique segment ID.

For segment analysis, true positives (\(TP\)) are defined as ground truth pixels that belong to a segment that overlaps with a predicted segment. False positives (\(FP\)) are predicted pixels whose segment does not overlap with a ground truth segment. False negatives (\(FN\)) are ground truth pixels whose segment does not overlap with a prediction segment. 

\begin{figure}[b!]
\centering
\includegraphics[width=1.0\columnwidth]{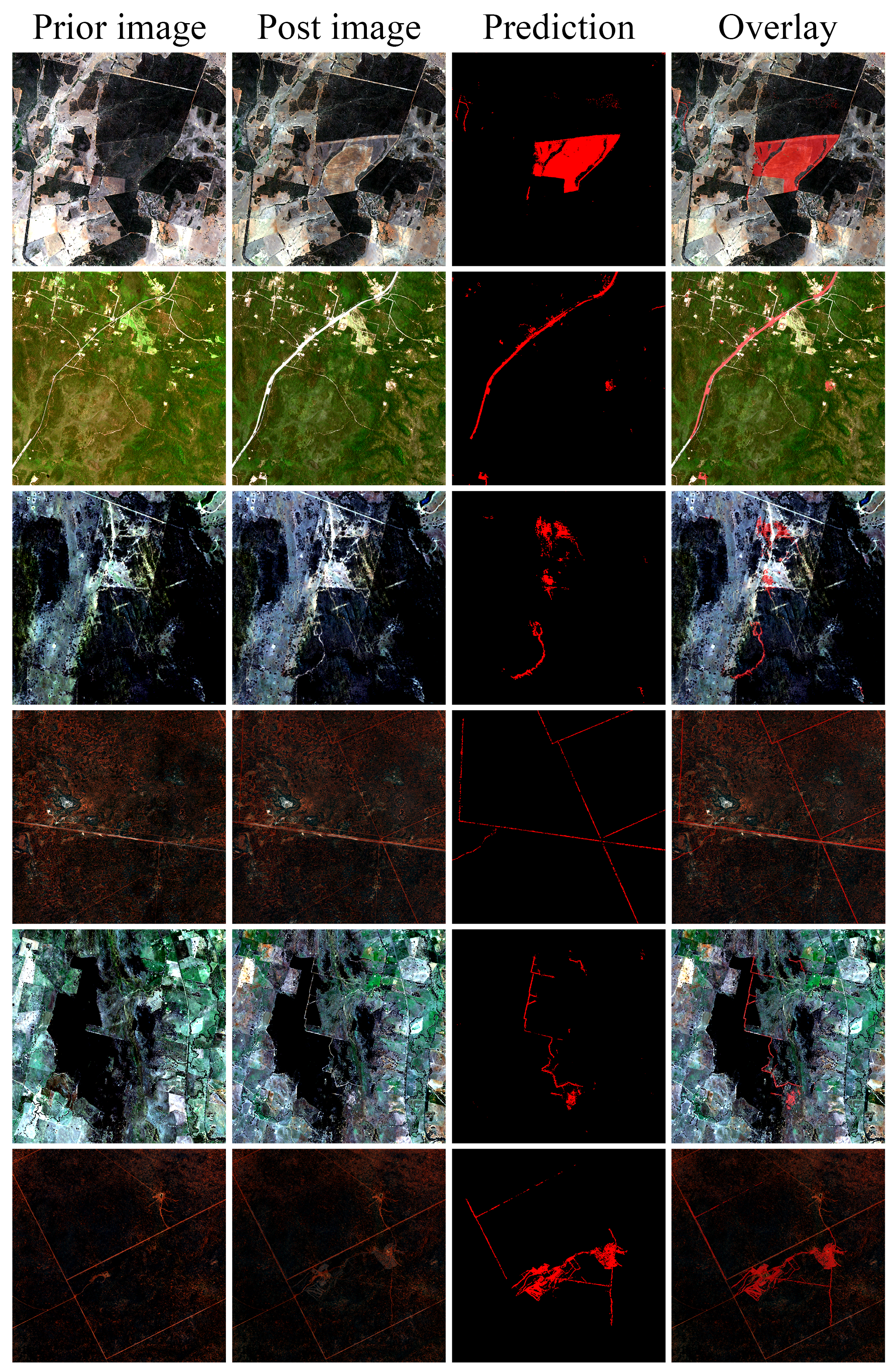}
\caption{Example predictions of the woody clearing model.}
\label{fig_clearingExample}
\end{figure}

Distance analysis aims to test the model’s ability to detect areas containing woody clearing and to assess the spatial error in the predictions. For distance analysis, a true positive (\(TP\)) is defined as a ground truth pixel that is within a threshold distance to any predicted woody clearing pixel. False positives (\(FP\)) are predicted pixels that are not within a threshold distance to any ground truth pixel. False negatives (\(FN\)) are ground truth pixels that are not within a threshold distance to any predicted pixel.

Example predictions of the trained woody clearing detection model can be seen in Fig.~\ref{fig_clearingExample}. 
Results of the woody clearing evaluation can be seen in section~\ref{section_clearingDetection}.

\subsection{Woody clearing loss scaling}
\subsubsection{Training}\label{section_lossScalingTraining}
To assess the model’s response to the loss scaling coefficient \(\alpha\) outlined in equations \ref{Equation_alphaDice} \& \ref{Equation_alphaBCE}, the model trained in section~\ref{section_clearingDetectionExp} was further fine tuned using various \(\alpha\) values ranging from \(\frac{1}{10}\) to \(10\). For each of the 15 selected \(\alpha\) values, the base model was trained for an additional 32 epochs (100,000 optimization iterations) using an identical methodology described in section~\ref{Section_clearingDetectionTraining}, with the exclusion of the total epochs trained for and loss scale \(\alpha\).

\subsubsection{Evaluation}
Each of the 15 models were evaluated on the (2023-2024) era following the same procedure outlined in section~\ref{section_clearingDetectionEval}. For each model, a confidence threshold value of 50\% was used to compute \(TP\), \(FP\) \& \(FN\). 
To better understand the relationship between precision and recall for various \(\alpha\) values, the \(F1\) and \(F2\) score can be generalized to an \(F_{\beta}\) score defined as:
\begin{equation}
F_\beta = \frac{(1 + \beta^2) \cdot \text{Precision} \cdot \text{Recall}}{\beta^2 \cdot \text{Precision} + \text{Recall}}
\end{equation} 
Where (\(\beta = 1\)) is equivalent to an \(F1\)-score which equally weights precision and recall, whilst for values (\(\beta > 1\)) recall is favored over precision and conversely for (\(\beta < 1\)) where precision is favored over recall. 

To establish a guideline for determining what \(\alpha\) value best optimizes for a specific target metric, a curve was fit to the observed precision and recall values for each \(\alpha\) conditioned model. The proposed models for estimating precision (\(\hat{P}\)) and recall (\(\hat{R}\)) with respect to \(\alpha\) are defined as:
\begin{equation}\label{eq_pHat}
\hat{P} = \text{sigmoid}(-\text{sign}(\alpha^\prime) \cdot \log_2(\sqrt{|\alpha^\prime|+1}) - \gamma_P)
\end{equation}
and
\begin{equation}\label{eq_rhat}
\hat{R} = \text{sigmoid}(\text{sign}(\alpha^\prime) \cdot \log_2(\sqrt{|\alpha^\prime|+1}) - \gamma_R).
\end{equation}
Where \(\alpha^\prime\) is a linearization transformation applied to \(\alpha\) defined as 
\begin{equation}
\alpha^\prime = 
\begin{cases} 
\alpha - 1,&  \text{if } \alpha > 1 \\
1 - \frac{1}{\alpha}, & \text{otherwise}.
\end{cases}
\end{equation}
\(\gamma_P\) and \(\gamma_R\) are curve origin offsets defined as:
\begin{equation}
\gamma_P = \log_e(\frac{1}{P_{\alpha1}} -1)
\end{equation}
and
\begin{equation}
\gamma_R = \log_e(\frac{1}{R_{\alpha1}} -1),
\end{equation}
where \(P_{\alpha1}\) and \(R_{\alpha1}\) denote the precision and recall scores for the default \(\alpha = 1\) model.

Results of the loss scaling experiment can be seen in section~\ref{section_lossScalingResults}.

\subsection{Woody clearing comparison}
To evaluate the proposed model’s ability to generalize to regions outside of the study area, a comparison study was performed using the MapBiomas \cite{souza2020reconstructing} dataset.
The MapBiomas \cite{souza2020reconstructing} project continuously maps vegetation, land use change and deforestation across Brazil, and provides date encoded polygons of deforestation alerts to signal regions experiencing woody clearing.  
A total of 6 Sentinel-2 scenes spread across Brazil were chosen for evaluation. Sentinel-2 image pairs were chosen based on the least cloudy image within a 3-month window from the end of 2022 and end of 2023 for the prior and post images respectively. Due to the absence of cloud free imagery, OmniCloudMask \cite{wright2025training} was used to generate cloud masks which excluded cloudy pixels in the prior or post image from statistical analysis.
Deforestation alert polygons were filtered to those spatially and temporally contained within the evaluation scenes’ prior and post images and were subsequently rasterized to overlay with the Sentintel-2 imagery. 
Model predictions for evaluation scenes were generated following the process outlined in section~\ref{section_clearingDetectionEval}

Prior work that maps woody vegetation clearing is the Global Forest Change \cite{hansen2013high} program, which provides annual global maps of tree clearing.
The Global Forest Change \cite{hansen2013high} clearing map was evaluated on the proposed validation dataset outlined in section~\ref{section_traingDataset} and the evaluation scenes used within the MapBiomas \cite{souza2020reconstructing} dataset.
The 30m Global Forest Change \cite{hansen2013high} clearing rasters corresponding to evaluation scenes were merged, clipped and projected to their respective scenes such that ground truth and Global Forest Change \cite{hansen2013high} predictions could be overlayed.
To account for temporal discrepancies between ground truth labels and predictions, clearing predictions within the current and prior annual time period were used to compute evaluation metrics.
The true positives for Global Forest Change \cite{hansen2013high} evaluation metrics used the sum of true positives contained within the annual clearing maps for both the current and prior time periods relative to the ground truth label.
The false positives and false negatives for Global Forest Change \cite{hansen2013high} evaluation metrics used only the current annual clearing map for the ground truth label, using the true positives to mask potential false positives or false negatives.
The approach ensures that the Global Forest Change \cite{hansen2013high} evaluation is not disadvantaged due to the ground truth label time period not aligning with Global Forest Change \cite{hansen2013high} annual reporting dates.

Results of the woody clearing comparison experiments can be seen in section~\ref{section_woodyClearingComparison}.

\subsection{Zero-shot woody segmentation}\label{section_woodyDetectionExperiment}
To assess the model’s ability to generalize to unseen tasks, the model was deployed in a zero-shot fashion to create binary segmentation masks of woody vegetation without any training through visual prompting. 
To extract the model’s internal representation of woody vegetation which is formulated when independently analysing the prior and post images for the woody clearing task, the prior image \(X_{t0}\) was given a reference image for which the woody segmentation mask should be generated for, whilst the post image \(X_{t1}\) was generated to extract the model’s latent representation of woody vegetation.

To assess the optimal post image generation technique to extract latent woody vegetation information from the model, an experiment was performed which benchmarked 9 different post image generation techniques and their efficacy in zero-shot woody segmentation. The 9 image generation techniques include: (1) zero image input, where the post image was replaced as an empty zero array. (2) Zero feature input, where all features (\(F_{t1}^{1}\), \(F_{t1}^{2}\), \(F_{t1}^{3}\), \(F_{t1}^{4}\), \(F_{t1}^{5}\)) from the encoder were replaced with zero arrays. Both image generation techniques (1) \& (2) aim to assess whether the absence of all features is sufficient to extract latent woody vegetation information from the network, or if additional context from the post image is required.

Image generation technique (3), empty scene, replaces the post image with an empty dirt patch which was tiled to fit the input resolution of 256w \(\times\) 256h. (4) Clearing scene, replaces the post image with a collection of clearing patches which were rotated, flipped and stitched together to fit the input resolution of 256w \(\times\) 256h. Image generation techniques (3) \& (4) both aim to test if features in the post image are necessary to extract woody information compared to the absence of all features in methods (1) \& (2). Method (4) tests whether features explicitly encoding clearing are required or if a generic scene with the absence of woody vegetation is sufficient as shown in method (3). Both image generation methods (3) \& (4) utilize image patch tiling due to difficulties of obtaining 2.56km \(\times\) 2.56km continuous regions of the desired features.

Image generation methods (5) \& (6) aim to augment the post images in methods (3) \& (4) respectively by replacing a 16-pixel wide border around the post image with the contents of the prior image. Image generation methods (7) \& (8) alternatively augment methods (3) \& (4) by pasting small, isolated patches of woody vegetation on top of the post images from a collection of 4 woody vegetation samples taken from a single scene. Methods (5) to (8) aim to assess whether additional contextual grounding of the post image is required by providing examples of woody vegetation to compare to. For methods (5) - (8), an additional weight mask was used to zero-weight predictions resultant from the border or from pasted woody vegetation samples. For methods (7) \& (8), a total of 8 post images were generated with different placement of pasted woody vegetation such that each pixel across the 8 generated images contained a non-zero weighting.

Image generation methods (1) – (8) all demonstrate methods utilizing handcrafted features, which requires the operator to have a deep understanding of the dataset and model to construct an appropriate post image. To propose a more generalizable post image generation method that is not dependent on the specific context of the task, a post image was generated using activation maximization optimization.

\begin{figure}
\centering
\includegraphics[width=0.95\columnwidth]{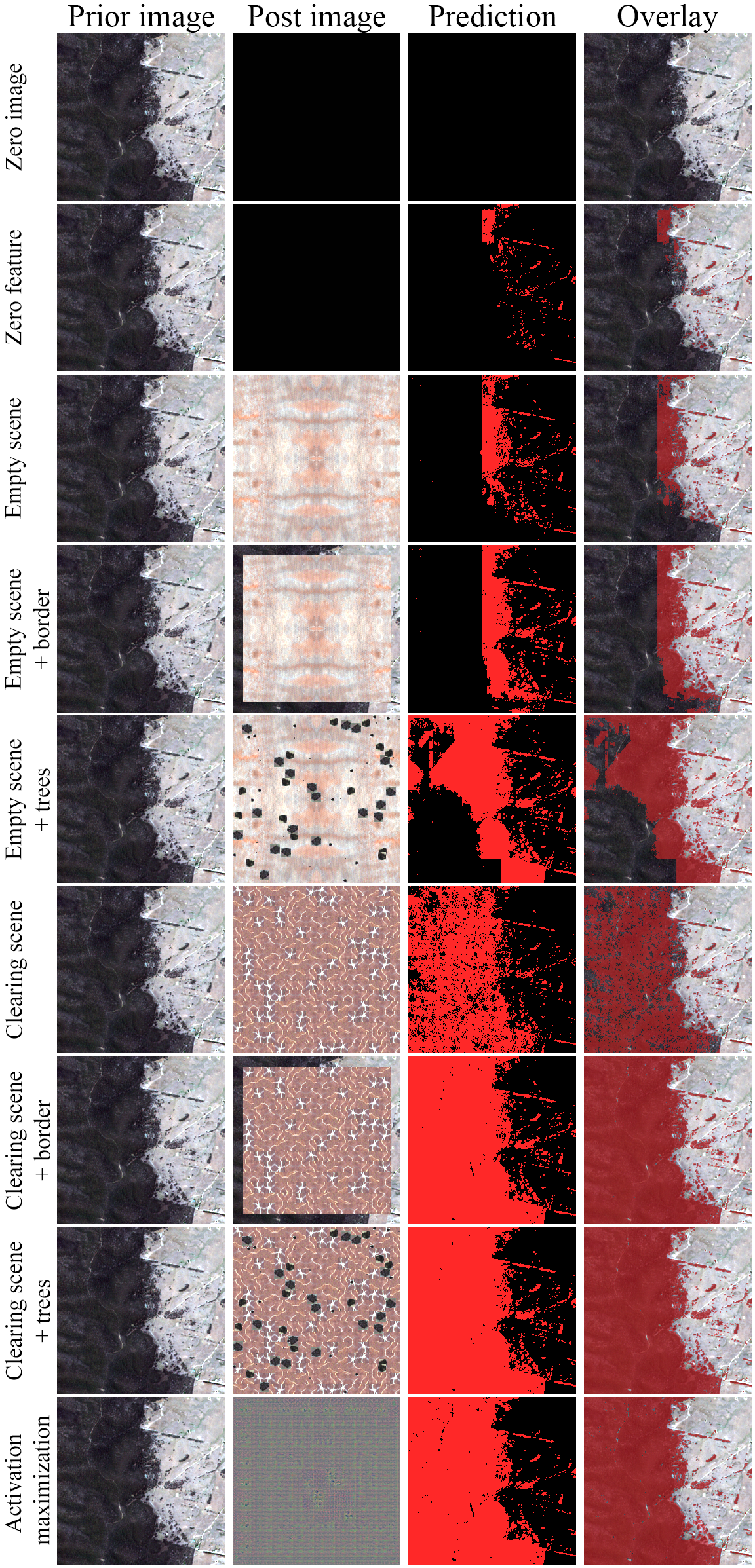}
\caption{Examples of the different post image generation techniques used to generate zero-shot woody predictions. Predictions were generated using a confidence threshold of 50\% using a masked weighted average and 50\% image overlap.}
\label{fig_woodyExample}
\end{figure}

Unlike traditional model training where the model’s weights are treated as parameters which are to be optimized to satisfy a loss function, images generated via activation maximization keep the model weights fixed and treat the input image as an array of parameters needing to be optimized. The objective of the optimization procedure is to find a post image \(X_{t1}\) that produces the maximum value of the model’s output prediction \(\hat{Y}_i\), which is achieved using the mean squared error between the prediction \(\hat{Y}_i\) and the largest confidence value the model can output:
\begin{equation}\label{Equation_activationLoss}
L_{Activation} = \frac{\sum\limits_{i=1}^{n} (\hat{Y}_i - 1.0)^{2}}{n}.
\end{equation}  
To constrain the optimization to prevent post images being generated containing large pixel intensities, a regularization term is introduced to the loss function which applies the mean squared error to the current generated image pixel intensities: 
\begin{equation}\label{Equation_inputLoss}
L_{Input} = \frac{\sum\limits_{i=1}^{n} (X_{t1i})^{2}}{n}.
\end{equation}     

A post image was generated using the input data tiles collected from the training dataset described in Section~\ref{section_traingDataset}. The input tiles were cropped to 256w \(\times\) 256h and fed into the model as the prior image, whilst the generated image was fed into the model as the post image using a batch size of 64. The loss and gradients were subsequently computed and the generated image’s pixel intensities were updated using stochastic gradient descent using an initial learning rate of 0.01 which was linearly decayed to 0.0001 across 20,000 optimization iterations.

Examples of the 9 proposed post image generation methods can be seen in Fig.~\ref{fig_woodyExample}.

\subsubsection{Zero-shot woody segmentation evaluation}

To assess the 9 proposed post image generation methods, a dataset of woody vegetation was created. 6 candidate scenes were chosen from spatially diverse regions across the state of New South Wales, Australia. From the 6 candidate scenes, randomly sampled patches were hand labeled denoting pixels containing woody vegetation and those that do not. Over 1M pixels were labeled, covering 108km\(^2\) of landmass where 28\% of pixels were considered woody vegetation.

Each of the 9 proposed post image generation methods were used to generate predictions of woody vegetation on the woody vegetation dataset using an input image size of 256w \(\times\) 256h. A masked weighted average was used to generate predictions where masked regions include the prior image border applied to the post image, and the pasted woody vegetation within the post image.
Predictions were generated using a 50\% image overlap where the border post image generation methods (5) and (6) used a 50\% image overlap of the valid unmasked region.

All handcrafted image generation approaches (1) – (8) used the default \(\alpha\) = 1 model to generate predictions. For image generation approach (9) activation maximization, due to the optimization process being explicitly aimed at generating a post image that denotes all pixels to be considered woody vegetation, using the default \(\alpha\) = 1 model would lead to small precision scores.
To counteract such adverse behaviour, the \(\alpha\) = \(\frac{1}{10}\) model which enhances precision at the cost of recall was used to attain a more even balance between precision and recall.

The models were evaluated under the metrics defined in \ref{section_clearingDetectionEval} with the exception of the F2 score being replaced with the overall accuracy metric (OA) which is defined as:
\begin{equation}\label{Equation_OA}
\text{OA} = \frac{TP + TN}{TP + FP + TN + FN}.  
\end{equation}
The F2 score is replaced by OA due to the woody clearing task being heavily class imbalanced, leading to non-meaningful high OA scores. As the woody segmentation dataset has a more even class balance and the woody segmentation task not focused on maximizing recall, the OA metric is favored.    

Results of the woody segmentation experiment can be seen in section~\ref{section_woodySegEval}.

\subsubsection{Zero-shot woody segmentation comparison}
To compare the proposed model’s zero-shot woody segmentation performance against prior work, SamGeo \cite{osco2023segment} was evaluated on the woody vegetation dataset. SamGeo’s \cite{osco2023segment} text-based pipeline utilizes GroundingDino \cite{liu2024grounding} for zero-shot visual grounding coupled with SAM \cite{kirillov2023segment} object segmentation.
Each image within the woody vegetation dataset was passed through the SamGeo’s \cite{osco2023segment} text-based pipeline with the prompt “Woody vegetation” to segment only woody vegetation.

Results of the zero-shot woody segmentation comparison can be seen in section~\ref{section_woodySegEval}.

\subsubsection{Zero-shot woody segmentation Fisher et al. point dataset}
To benchmark the performance of the model’s ability to zero-shot the woody segmentation task in relation to prior works, an experiment was performed using the labelled point dataset from \cite{fisher2016large}. The \cite{fisher2016large} Fisher et al. dataset contained 6,648 points sampled using a stratified random approach across the state of New South Wales, Australia where each point was manually classified through visual interpretation using a Leica ADS40 airborne digital camera at 0.5m resolution captured in 2011.

Each of the 9 proposed post image generation methods were used to generate a complete woody vegetation map of New South Wales, Australia, covering 801,137 km\(^2\) using the 2018 Sentinel-2 imagery. Each of the generation methods alongside the \cite{fisher2016large} Fisher et al. model were evaluated under 3 criteria: woody accuracy, non-woody accuracy and overall accuracy defined as: 
\begin{equation}\label{Eq_WoodyAccuracy}
\text{woody accuracy} = \frac{\sum\limits_{i=1}^{n}{\hat{Y}_i Y_i}}{\sum\limits_{i=1}^{n}{Y_i}},
\end{equation}
\begin{equation}\label{Eq_NonWoodyAccuracy}
\text{non-woody accuracy} = \frac{\sum\limits_{i=1}^{n}{(1 - \hat{Y}_i) (1 - Y_i)}}{\sum\limits_{i=1}^{n}{(1 - Y}_i)}
\end{equation}
and
\begin{equation}\label{Eq_OverallAccuracy}
\text{overall accuracy} = \frac{\sum\limits_{i=1}^{n}{\hat{Y}_i Y_i} + \sum\limits_{i=1}^{n}{(1 - \hat{Y}_i) (1 - Y}_i)}{n},
\end{equation}
where \(\hat{Y}_i\) and \(Y_i\) denote the binary prediction of the model using a 50\% confidence threshold and the ground truth label respectively.

The results of the zero-shot woody segmentation on the \cite{fisher2016large} Fisher et al. dataset can be seen in Section~\ref{section_woodySegFisher}.

\subsection{Zero-shot woody regrowth detection}
To further assess the model’s ability to perform binary zero-shot segmentation, the model was deployed with the objective to detect woody regrowth. For this work, binary woody regrowth is defined as pixels in the present that satisfy the woody definition criteria of vegetation over the height of 2m, which are resultant from clearing events in the past. To extract woody regrowth information from the network, the input pairs \(X_{t0}\) and \(X_{t1}\) were swapped such that the original prior image became the post image and the original post image became the prior image. These switched input pairs are defined as the pseudo-prior and pseudo-post images.

To assess the model’s ability to zero-shot woody regrowth detection, an evaluation dataset was created. 
Unlike woody clearing, woody regrowth is visually subtle in appearance due to being a continuous process occurring over several years. In contrast to woody clearing which is derived from discrete events occurring on a specific date.
Therefore to provide sufficient time for woody regrowth to occur, the pseudo-prior images were taken from the Sentinel-2 images for the year 2024, whilst the pseudo-post images were taken from 2018. 

Despite the rarity of woody clearing, woody regrowth is even less prevalent to observe. Combined with the ambiguity of its non-discrete appearance, it becomes difficult to randomly assign regions for manual labeling due to the low probability of overlap and the difficulty for humans to naturally detect it whilst scanning the image.   
Therefore to systematically propose regions for human labeling, the woody clearing model trained with loss scaling coefficient \(\alpha = 4\) was used to propose scenes with high woody regrowth present.
The \(\alpha = 4\) model was chosen as an initial region proposal method due to its higher recall scores compared to the standard \(\alpha = 1\) model, making it more suitable for alerting operators which scenes to investigate for manual labeling. 

A total of 5 scenes across New South Wales, Australia were proposed. Manual labeling focused on regions within the scene that were flagged by the model to contain woody regrowth.
Over 490,000 pixels were labeled, covering 49km\(^2\) of landmass where 16\% of pixels were considered woody regrowth.

\begin{figure}
\centering
\includegraphics[width=1.00\columnwidth]{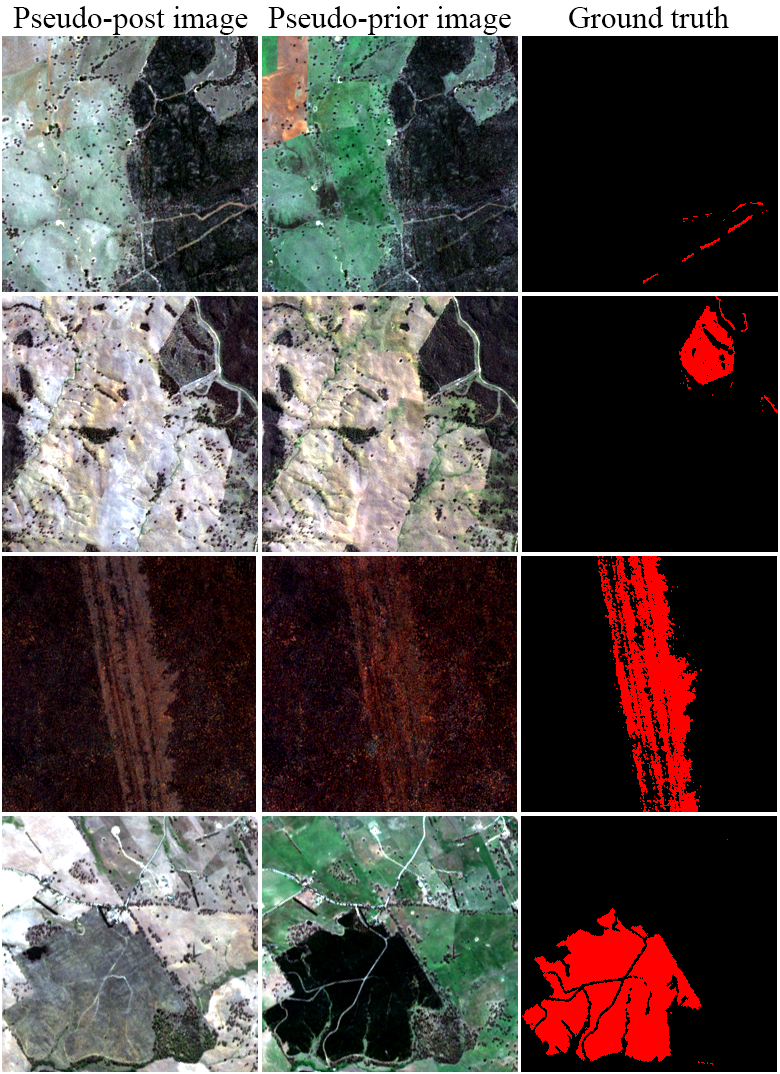}
\caption{Example images used to evaluate woody regrowth, where the pseudo-post image derives from 2018 Sentinel-2 imagery whilst the pseudo-prior image derives from 2024 Sentinel-2 imagery.}
\label{fig_regrowthExample}
\end{figure}

Examples of the woody regrowth dataset can be seen in Fig.~\ref{fig_regrowthExample}.
Results of the zero-shot woody regrowth detection experiment can be seen in section \ref{section_regrowthResults}.

\subsubsection{Zero-shot woody regrowth comparison}
To compare the proposed model’s zero-shot woody regrowth segmentation performance against prior work, Segment Any Change \cite{zheng2024segment} was evaluated on the woody regrowth dataset. Segment Any Change \cite{zheng2024segment} utilizes SAM \cite{kirillov2023segment} latent vectors from bitemporal image pairs to detect semantic changes in input imagery. Change predictions for each image within the woody regrowth dataset were generated for evaluation.

To derive additional insights into how much spatial contextual information the proposed approach utilizes when performing zero-shot woody regrowth segmentation, a rudimentary NDVI difference threshold baseline was proposed to compute the reliance on individual pixel spectral values.
For each image pair within the woody regrowth dataset, the NDVI vegetation index for each pixel in each image was computed as:
\begin{equation}
NDVI = \frac{{NIR} – {R}}{{NIR} + {R}},
\end{equation}
where a binary regrowth prediction for each pixel was determined by thresholding the NDVI difference between the post and prior images:
\begin{equation}
\hat{Y} = ({NDVI}_{post} – {NDVI}_{prior})  > {Threshold}.
\end{equation}
To determine a suitable threshold to apply for an image pair, threshold values beginning from 0.0 to 2.0 using increments of 0.1 were applied to all images excluding the target image pair. From the threshold values, the threshold corresponding to the highest F1 score was subsequently applied to the excluded target image pair. The process of excluding a new target image pair was repeated until all image pairs had been assigned a threshold.

The results of the zero-shot regrowth comparison experiments can be seen in section \ref{section_regrowthResults}.


\section{Results}
\subsection{Woody clearing detection}\label{section_clearingDetection}
The results for woody clearing detection are shown in table~\ref{table_metrics} at confidence thresholds ranging from 10\% to 90\%.
The model attains a precision of 35.2\% with a recall of 79.7\% at a confidence level of 50\% on a pixel-wise metric basis. 

When comparing the pixel-wise results to the segment-wise results it can be observed that precision and recall increase to 56.9\% and 97.5\% respectively, resulting in an increase of precision by 1.6\(\times\) and a reduction in missed detections by a factor of 8.
This large decrease in missed detections indicates that the model excels at proposing regions where woody clearing is present, but experiences difficulties in precisely matching individual pixel predictions to human labels.

This is further supported by observing the recall metric across different distance thresholds as shown in table~\ref{table_distance}. By applying a 10m distance threshold which equates to a 1-pixel buffer, the amount of missed woody clearing by the model reduces by a factor of 1.6 and is further reduced by a factor of 1.9 when applying a 20m distance threshold.

The disparity between the model’s achieved precision and recall metric is further discussed in section~\ref{section_precisionDiscussion}.

\begin{table}
\centering
\caption{Performance metrics for woody clearing detection on a pixel-wise and segment-wise basis}\label{table_metrics}
\begin{tabular}{lccccc}
\toprule
Confidence & Precision & Recall & F1 & F2 & IoU \\
\midrule
\multicolumn{6}{c}{Pixel-wise metrics} \\
\hline
10\% & 0.289 & 0.830 & 0.429 & 0.604 & 0.273 \\
25\% & 0.312 & 0.819 & 0.452 & 0.618 & 0.292 \\
50\% & 0.352 & 0.797 & 0.488 & 0.636 & 0.323 \\
75\% & 0.390 & 0.775 & 0.519 & 0.647 & 0.351 \\
90\% & 0.409 & 0.763 & 0.533 & 0.651 & 0.363 \\
\hline
\multicolumn{6}{c}{Segment-wise metrics} \\
\hline
10\% & 0.463 & 0.983 & 0.630 & 0.803 & 0.459 \\
25\% & 0.503 & 0.980 & 0.665 & 0.824 & 0.498 \\
50\% & 0.569 & 0.975 & 0.718 & 0.853 & 0.561 \\
75\% & 0.626 & 0.970 & 0.761 & 0.874 & 0.614 \\
90\% & 0.653 & 0.967 & 0.780 & 0.882 & 0.639 \\
\bottomrule
\end{tabular}
\end{table}

\begin{table}
\centering
\caption{Performance metrics for woody clearing detection using distance thresholds}\label{table_distance}
\begin{tabular}{lccccc}
\toprule
Distance & Precision & Recall & F1 & F2 & IoU \\
\midrule
0m & 0.352 & 0.797 & 0.488 & 0.636 & 0.323 \\
10m & 0.451 & 0.871 & 0.594 & 0.734 & 0.423 \\
20m & 0.478 & 0.891 & 0.622 & 0.760 & 0.452 \\
30m & 0.492 & 0.901 & 0.636 & 0.772 & 0.467 \\
40m & 0.503 & 0.909 & 0.648 & 0.783 & 0.479 \\
50m & 0.510 & 0.914 & 0.655 & 0.789 & 0.487 \\
60m & 0.515 & 0.918 & 0.660 & 0.794 & 0.493 \\
70m & 0.520 & 0.923 & 0.665 & 0.799 & 0.498 \\
80m & 0.524 & 0.926 & 0.670 & 0.803 & 0.503 \\
90m & 0.528 & 0.930 & 0.674 & 0.807 & 0.508 \\
100m & 0.531 & 0.932 & 0.677 & 0.810 & 0.511 \\
\bottomrule
\end{tabular}
\\ \vspace{1pt}
\raggedright
Metrics were computed under a confidence threshold of 50\%.
Spatial resolution of input imagery was 10m per-pixel.
\end{table}

\subsection{Woody clearing loss scaling}\label{section_lossScalingResults}
\begin{figure}
\centering
\includegraphics[width=1.00\columnwidth]{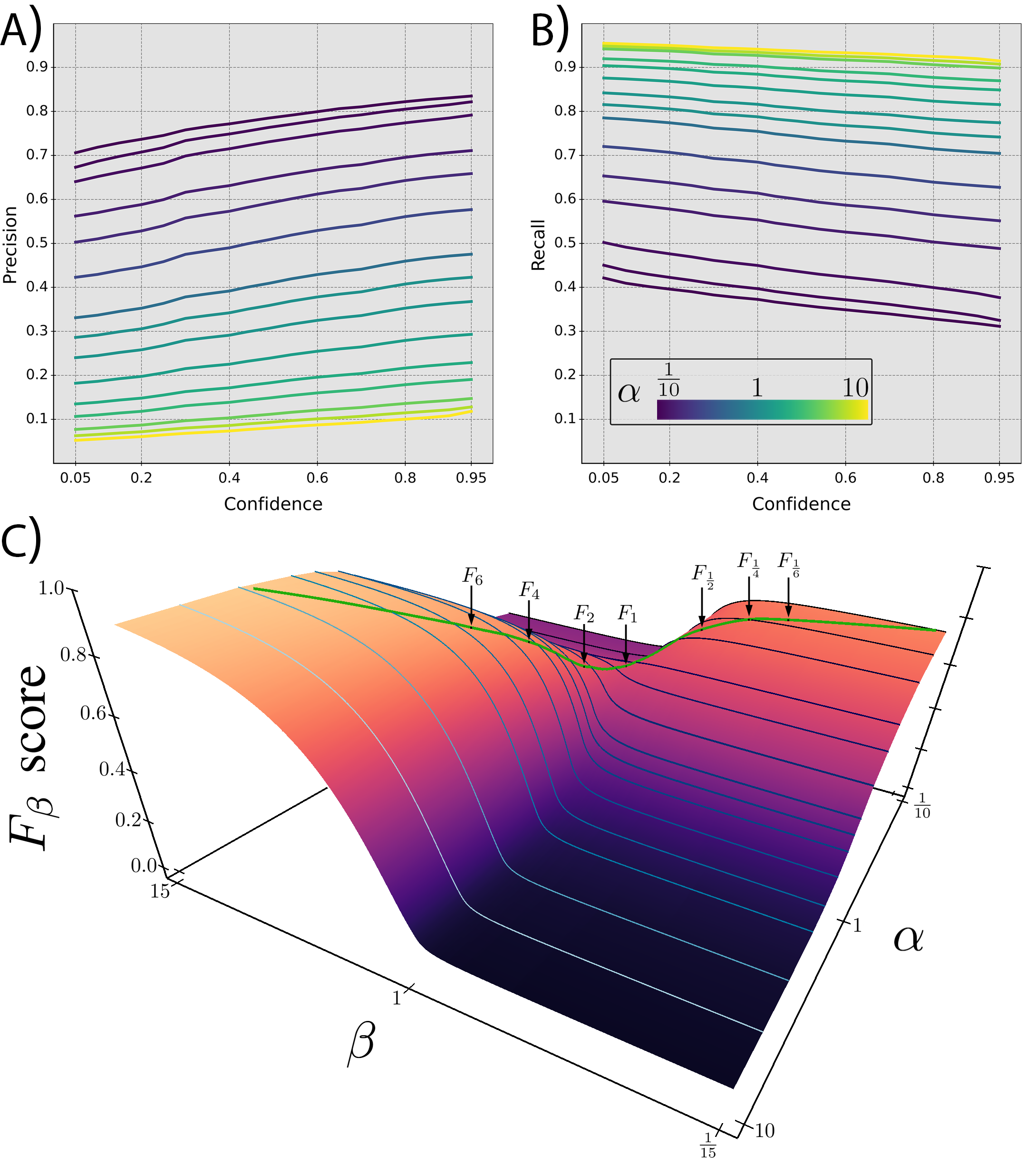}
\caption{Results from the woody clearing loss scaling experiment.
Figures A) and B) plot the attained precision and recall metrics across confidence thresholds ranging from 5\% to 95\% against various \(\alpha\) conditioned models where \(\alpha\) ranges from \(\frac{1}{10}\) to 10. 
Figure C) plots the \(F_{\beta}\) scores attained for each \(\alpha\) conditioned model where \(\beta\) ranges from \(\frac{1}{15}\) to 15. The green line indicates the maximum \(F_{\beta}\) score for a specific \(\beta\) value, highlighting the optimal \(\alpha\) value which maximizes the particular \(F_{\beta}\) score.}
\label{alphaScaling}
\end{figure}

The results of applying the loss scaling coefficient \(\alpha\) with respect to model performance can be seen in Fig.~\ref{alphaScaling}. Observing Fig.~\ref{alphaScaling}-A \& Fig.~\ref{alphaScaling}-B, as the loss scaling coefficient \(\alpha\) increases, the model attains a greater recall at the cost of reduced precision. Whilst the inverse is achieved by reducing \(\alpha\) to boost precision at the expense of recall.
However observing both Fig.~\ref{alphaScaling}-A \& Fig.~\ref{alphaScaling}-B it can be seen that a greater overall recall or precision can be attained by optimizing the loss scaling coefficient \(\alpha\) compared to traditional confidence thresholding. A full table of precision and recall scores conditional on \(\alpha\) and confidence thresholds can be seen in tables \ref{table_alphaConfPrecision} and \ref{table_alphaConfRecall} in appendix~\ref{appendix_alphaConf}.

\begin{figure}
\centering
\includegraphics[width=1.0\columnwidth]{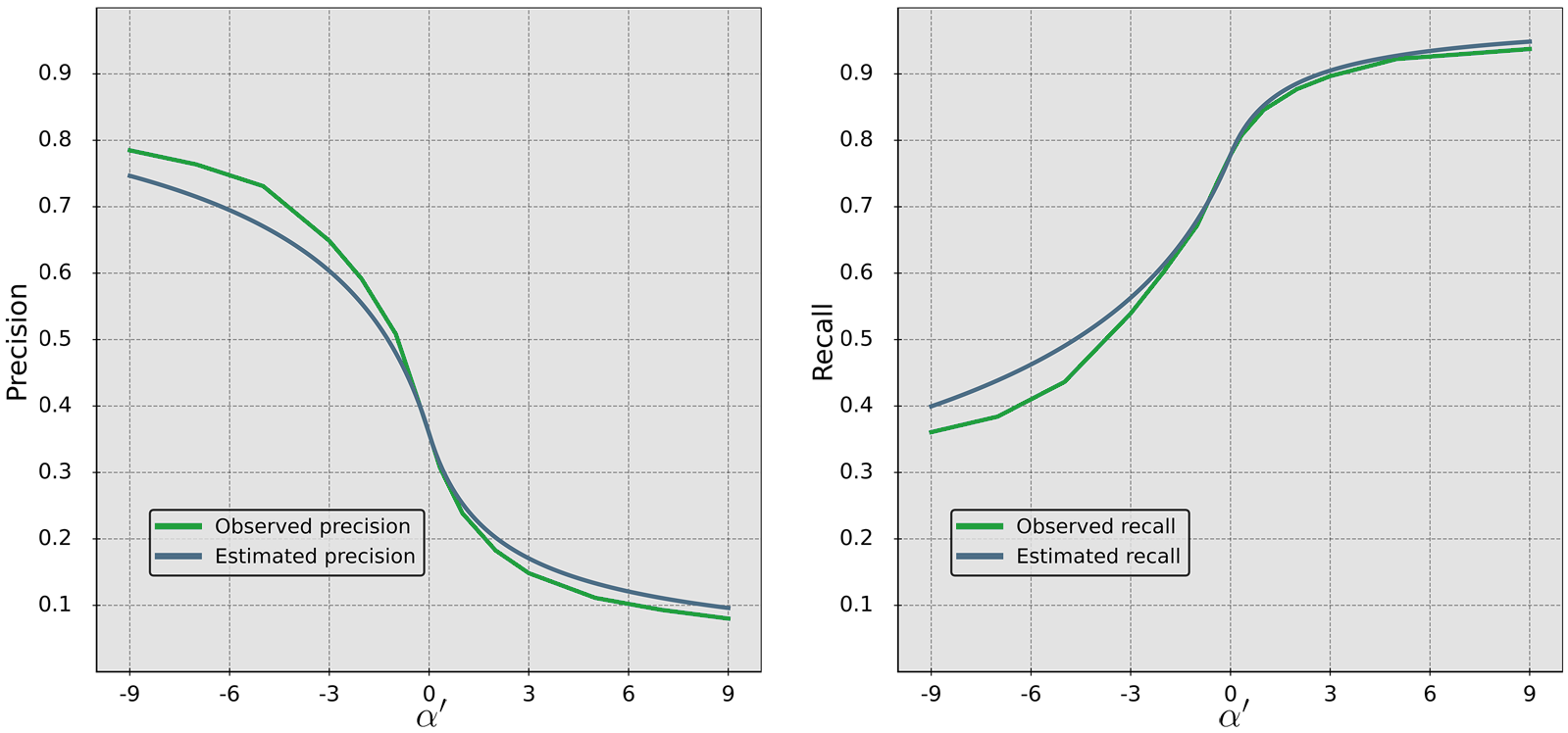}
\caption{Plots comparing the observed precision and recall metrics for the woody clearing detection task to that of the estimated precision and recall using equations \ref{eq_pHat} and \ref{eq_rhat}}
\label{fig_alphaPrecisionRecallRegression}
\end{figure}

Each model trained on a given \(\alpha\) parameter was assessed on various \(F_{\beta}\) scores where \(\beta\) ranges from [\(\frac{1}{15}\), \(15\)] which can be seen in Fig.~\ref{alphaScaling}-C. 
It can be see that as \(\beta\) increases, the relative performance of models with greater \(\alpha\) values also increases. This is due to higher \(\beta\) values more heavily weight recall when assessing model performance which higher \(\alpha\) models optimize for. 
However given a specific target metric e.g. \(F_4\) score, there exists an optimal \(\alpha\) value that optimizes for the target metric. The data derived optimal \(\alpha\) values for given \(\beta\) values are shown as the green curve in Fig.~\ref{alphaScaling}-C.

A comparison between the observed precision and recall scores attained by the \(\alpha\) conditioned models and the estimated scores via equations \ref{eq_pHat} and \ref{eq_rhat} can be seen in Fig.~\ref{fig_alphaPrecisionRecallRegression}. Terms \(P_{\alpha1}\) and \(R_{\alpha1}\) are assigned a value of 0.352 and 0.797 respectively and are derived from table~\ref{table_metrics}.

From Fig.~\ref{fig_alphaPrecisionRecallRegression}, it can be observed that the curves estimating the precision and recall scores closely align to the shape of the observed values. The average error between each of the observed points and predicted points for precision and recall are 0.025 and 0.016 respectively, where the majority of the error deriving from lower value \(\alpha^\prime\) estimates.

Additional plots comparing observed precision and recall scores against predicted precision and recall scores for the zero-shot woody segmentation and regrowth detection tasks can be seen in Fig.~\ref{fig_alphaPrecisionRecallRegression_woody} and Fig.~\ref{fig_alphaPrecisionRecallRegression_regrowth} in appendix~\ref{appendix_precisionRecallEstimation}.
Further plots comparing the observed precision and recall scores against alternative scaled loss functions can be found in Fig~\ref{fig_alphaPrecisionRecall_Tversky} in appendix~\ref{appendix_scaledLossComparison}.

\subsection{Woody clearing comparison}\label{section_woodyClearingComparison}

The results of the proposed model and Global Forest Change \cite{hansen2013high} on the MapBiomas \cite{souza2020reconstructing} dataset can be seen in table~\ref{table_clearingComparison}.

\begin{table}[b]
\centering
\caption{Woody clearing performance on MapBiomas dataset}\label{table_clearingComparison}
\begin{tabular}{lccccc}
\toprule
Model & Precision & Recall & F1 & F2 & IoU \\
\midrule
Proposed approach & \textbf{0.166} & \textbf{0.735} & \textbf{0.270} & \textbf{0.436} &\textbf{0.156} \\
Global Forest Change & 0.084 & 0.494 & 0.144 & 0.251 & 0.078 \\
\bottomrule
\end{tabular}
\\ \vspace{1pt}
\raggedright
Bold values indicate the most optimal value for a given metric.
\end{table}

 The proposed model achieved a recall score of 0.735, a 1.5\(\times\) improvement compared to the Global Forest Change \cite{hansen2013high} recall score of 0.494. However, both models attained relatively low precision scores of 0.166 and 0.084, reducing the F1 scores of the proposed model and Global Forest Change \cite{hansen2013high} to 0.270 and 0.144 respectively. The lower precision score is attributable to missed clearing events within the MapBiomas \cite{souza2020reconstructing} dataset, where the majority of false positives from both the proposed model and Global Forest Change \cite{hansen2013high} were the result of correct detections of clearing events being mislabeled.

Global Forest Change \cite{hansen2013high} attained a precision, recall and F1 score of 0.158, 0.268 and 0.199 respectively across the New South Wales study area. Compared to the proposed model which achieved a 2.2\(\times\), 3.0\(\times\) and 2.45\(\times\) increased performance in precision, recall and F1 scores respectively.

A comparison of generated predictions and missed clearing events on the MapBiomas \cite{souza2020reconstructing} dataset can be seen in Fig.~\ref{fig_clearing_comparison} in section~\ref{section_modelComparison}.

\subsection{Zero-shot woody segmentation}\label{section_woodySegResults}
\subsubsection{Zero-shot woody segmentation evaluation}\label{section_woodySegEval}
The model’s performance using different post image generation techniques for the zero-shot woody segmentation task can be seen in table~\ref{table_woodyMetrics}. The greatest performing model was the clearing scene + trees post image generation method with an F1 score of 0.899. Handcrafted post image generation techniques that used real reference scenes performed greater than those that used zeroed inputs. On average, the post image generation techniques utilizing a clearing scene as reference attained an F1 score twice as great compared to those using the empty scene. Adding additional contextual woody vegetation further improved model performance, whereby randomly placing trees across the post image was shown to increase the base clearing scene F1 performance by a factor of 2.

Compared to the automated post image generation technique activation maximization, the best performing handcrafted post image generation method shared comparable performance to each other. The best handcrafted approached outperformed the activation maximization approach by 1.6 percentage points in terms of the overall accuracy.

Further insights into the post image generation techniques and their shared difficulties are discussed in section \ref{section_woodyDifficulties}. 

Compared to alternative works, SamGeo \cite{osco2023segment} achieved an F1 score of 0.685, where the best performing zero-shot configuration outperformed it by a factor of 1.3\(\times\). Comparison of generated predictions between SamGeo \cite{osco2023segment} and the Clearing scene + trees method can be seen in Fig.~\ref{fig_woody_comparison} in section~\ref{section_modelComparison}.

\begin{table}
\centering
\caption{Performance metrics on zero-shot woody segmentation}\label{table_woodyMetrics}
\begin{tabular}{lccccc}
\toprule
Model & Precision & Recall & F1 & IoU & OA \\
\midrule
Zero image & 0.829 & 0.000 & 0.000 & 0.000 & 0.723 \\
Zero feature & 0.666 & 0.058 & 0.106 & 0.056 & 0.731 \\
Empty-scene & 0.591 & 0.136 & 0.221 & 0.124 & 0.734 \\
Empty scene + border & 0.587 & 0.189 & 0.286 & 0.167 & 0.738 \\
Empty scene + trees & 0.684 & 0.375 & 0.485 & 0.320 & 0.779 \\
Clearing scene & \textbf{0.921} & 0.298 & 0.451 & 0.291 & 0.798 \\
Clearing scene + border & 0.874 & 0.524 & 0.655 & 0.487 & 0.847 \\
Clearing scene + trees & 0.904 & \textbf{0.895} & \textbf{0.899} & \textbf{0.817} & \textbf{0.944} \\
Activation maximization & 0.867 & 0.867 & 0.867 & 0.764 & 0.928 \\
\bottomrule
SamGeo & 0.683 & 0.688 & 0.685 & 0.521 & 0.792 \\
\bottomrule
\end{tabular}
\\ \vspace{1pt}
\raggedright
Metrics were computed under a confidence threshold of 50\%.
Bold values indicate the most optimal value for the given metric.
\end{table}

\subsubsection{Zero-shot woody segmentation Fisher et al. point dataset}\label{section_woodySegFisher}
The results of the model’s zero-shot woody segmentation performance on the \cite{fisher2016large} Fisher et al. dataset can be seen in table~\ref{table_fisher}. Similar to the zero-shot woody segmentation results in Section~\ref{section_woodySegEval}, the post image generation techniques which utilized a clearing scene and those that added additional contextual woody vegetation performed better than those which did not.

\begin{table}
\centering
\caption{Zero-shot woody segmentation performance metrics on Fisher et al. point dataset}\label{table_fisher}
\begin{tabular}{lccc}
\toprule
\multirow{2}{*}{Model} & Woody & Non-woody & Overall \\
 & accuracy & accuracy & accuracy \\
\midrule
Zero image & 0.000 & \textbf{1.000} & 0.536 \\
Zero feature & 0.132 & 0.978 & 0.585 \\
Empty-scene & 0.220 & 0.959 & 0.616 \\
Empty scene + border & 0.356 & 0.910 & 0.653 \\
Empty scene + trees & 0.468 & 0.891 & 0.695 \\
Clearing scene & 0.295 & 0.974 & 0.659 \\
Clearing scene + border & 0.756 & 0.930 & 0.849 \\
Clearing scene + trees & \textbf{0.883} & 0.899 & \textbf{0.892} \\
Activation maximization & 0.846 & 0.900 & 0.873 \\
\bottomrule
Fisher et al. & 0.747 & 0.930 & 0.868 \\
\bottomrule
\end{tabular}
\\ \vspace{1pt}
\raggedright
Metrics were computed under a confidence threshold of 50\%.
Bold values indicate the most optimal value for the given metric.
\end{table}

Relative to the \cite{fisher2016large} Fisher et al. model which was explicitly trained to detect woody vegetation, the proposed model using the post image generation techniques; activation maximization and clearing scene with isolated tree images, outperformed the \cite{fisher2016large} Fisher et al. model despite being zero-shot transferred. The \cite{fisher2016large} Fisher et al. model attained an overall accuracy of 86.8\%, whilst the zero-shot woody segmentation model achieved an overall accuracy of 87.3\% using activation maximization and 89.2\% using the clearing scene with trees, a reduction in the error by 3.8\% and 18.2\% respectively.

\subsection{Zero-shot woody regrowth detection}\label{section_regrowthResults}
The model’s performance for the zero-shot woody regrowth segmentation task can be seen in table~\ref{table_regrowthMetrics}. At a confidence threshold level of 50\% the model attains an F1 score of 0.845 and an overall accuracy of 94.7\%.

\begin{table}
\centering
\caption{Performance metrics on zero-shot woody regrowth}\label{table_regrowthMetrics}
\begin{tabular}{lccccc}
\toprule
Model & Precision & Recall & F1 & IoU & OA \\
\midrule
Proposed approach & 0.815 & 0.878 & 0.845 & 0.731 & 0.947 \\
Segment Any Change & 0.474 & 0.322 & 0.383 & 0.237 & 0.828 \\
NDVI threshold & 0.659 & 0.556 & 0.603 & 0.432 & 0.878 \\ 
\bottomrule
\end{tabular}
\raggedright
Bold values indicate the most optimal value for a given metric.
\end{table}

Compared to Segment Any Change \cite{zheng2024segment} and NDVI difference thresholding which attained an F1 score of 0.383 and 0.603 respectively, the proposed approach achieved a performance improvement by a factor of 2.2\(\times\) and 1.4\(\times\) respectively. Comparison of generated predictions between Segment Any Change \cite{zheng2024segment} and the NDVI difference thresholding method can be seen in Fig.~\ref{fig_regrowth_comparison} in section~\ref{section_modelComparison}.


\section{Discussion}
\subsection{Woody clearing precision performance}\label{section_precisionDiscussion}
When comparing precision and recall for the woody clearing detection task, table~\ref{table_metrics} shows a large disparity between both metrics' performances. It was found that recall was greater than precision by a factor of 2.26 at a 50\% confidence threshold, without applying any \(\alpha\) scaling to bias the model towards either metric.  
Observing the precision and recall scores for the zero-shot woody segmentation and woody regrowth tasks in tables~\ref{table_woodyMetrics} \& \ref{table_regrowthMetrics}, the large disparity between precision and recall does not exist, despite using identical model weights.

The cause of the poor performance of the precision metric in the woody clearing detection task is due to discrepancies of what is required to be reported as woody clearing when formulating the dataset's labels.
Woody clearing resultant from plantations were not required to be reported on and therefore were not present in the dataset’s labels. As plantations consist of large dense examples of woody clearing that comprise of many pixels, the model’s precision score is adversely affected due to any predicted clearing being incorrectly flagged as a false positive inside of plantations.

An additional factor that negatively impacts precision is the requirement that operators are only required to validate pixels derived from the initial woody clearing index. Despite the clearing index being calibrated to have a high recall at the cost of precision, small traces of false negatives can be found which causes operators to miss examples of woody clearing when formulating labels.

This discrepancy in what is reported as woody clearing from an operational perspective is further demonstrated in the MapBiomas \cite{souza2020reconstructing} dataset. Both the proposed work and Global Forest Change \cite{hansen2013high} attained low precision scores within the MapBiomas \cite{souza2020reconstructing} dataset due to non-reported clearing events deriving from minimum reporting area requirements, as seen in Fig.~\ref{fig_clearing_comparison}.

Compared to the woody segmentation and woody regrowth datasets, such semantic exclusion criteria from ignoring plantations do not exist and therefore present a more homogeneous performance between precision and recall.

\subsection{\(\alpha\) loss scaling}
Conditioning the binary cross-entropy and dice loss on the loss scaling coefficient \(\alpha\) demonstrated to be an effective method for targeting the optimization of specific \(F_{\beta}\) scores. The ability to train a model that maximizes recall for a given minimal precision threshold allows for a remote sensing product aimed towards end user specifications. 

Applying confidence thresholding alone was able to attain a precision score of 0.409 at 90\% confidence threshold and a recall score of 0.830 at 10\% confidence threshold as shown in table~\ref{table_metrics}. 
Further extreme confidence thresholding resulted in a precision score of 0.421 at 99.5\% confidence threshold and a recall score of 0.838 at 0.5\% confidence threshold, an increase in the precision and recall with respect to the 90\% and 10\% confidence thresholds of 2.9\% and 0.9\% respectively. Thus demonstrating confidence thresholding alone being an ineffective method to tune precision and recall scores as fine numerical precision is required due to deep learning models producing overconfident predictions.

By applying \(\alpha\) loss scaling, precision was able to be increased from 0.359 at an \(\alpha\) value of 1 and 50\% confidence threshold to 0.785 at an \(\alpha\) value of \(\frac{1}{10}\) and 50\% confidence threshold.
Recall was able to be increased from 0.778 at an \(\alpha\) value of 1 and 50\% confidence threshold to 0.937 at an \(\alpha\) value of 10 and 50\% confidence threshold.
Resulting in an increase in the default precision and recall scores of 118.7\% and 20.4\% respectively at a 50\% confidence threshold.
Compared to confidence thresholding alone, introducing the \(\alpha\) loss scaling coefficient was able to result in a model with a 1.85\(\times\) higher precision or 1.12\(\times\) higher recall.

\(\alpha\) loss scaling was further found to be useful during zero-shot woody segmentation where the activation maximization post image generation technique alone resulted in a precision and recall score of 49.2\% and 99.1\% respectively when using the \(\alpha\) = 1 model. By counteracting the activation maximization optimization objective of prioritising high recall scores with the prioritisation of precision scores for \(\alpha < 1\) models, the \(\alpha = \frac{1}{10}\) model achieved a precision and recall score of 86.7\%, a \(1.32\times\) increase in the F1 score when compared to the default \(\alpha\) = 1 model.

However to align the model’s objective with end user’s objectives, an appropriate \(\alpha\) value must first be estimated to condition the training process. Conditioning the model to optimize for a specific \(F_{\beta}\) score is dependent on the task and the dataset the model is deployed on as seen in Fig. \ref{fig_alphaPrecisionRecallRegression}, \ref{fig_alphaPrecisionRecallRegression_woody} \& \ref{fig_alphaPrecisionRecallRegression_regrowth}.
Equations \ref{eq_pHat} \& \ref{eq_rhat} demonstrate the general relationship between model performance and \(\alpha\) where an initial precision and recall score on a validation dataset is required at the origin \(\alpha = 1\) to calibrate for the specific task and dataset difficulty.
An \(\alpha\) value can be estimated to achieve a specific precision or recall score by solving for \(\alpha^\prime\)  in equations \ref{eq_pHat} \& \ref{eq_rhat} which results in:
\begin{equation}\label{eq_palpha}
\alpha^{\prime}_{P} =
\begin{cases} 
2^{2 \log_e(\frac{1 - \hat{P}}{\hat{P}}) – 2\gamma_P} - 1  &\text{if \(\alpha^{\prime}_{P} >0\)} \\
-2^{-2 \log_e(\frac{1 - \hat{P}}{\hat{P}}) + 2\gamma_P} + 1  &\text{else}
\end{cases}
\end{equation}
and
\begin{equation}\label{eq_ralpha}
\alpha^{\prime}_{R} =
\begin{cases} 
2^{-2 \log_e(\frac{1 - \hat{R}}{\hat{R}}) + 2\gamma_R} - 1  &\text{if \(\alpha^{\prime}_{R} >0\)} \\
-2^{2 \log_e(\frac{1 - \hat{R}}{\hat{R}}) – 2\gamma_R} + 1  &\text{else}
\end{cases},
\end{equation}
where \(\alpha^{\prime}_{P}\) \& \(\alpha^{\prime}_{R}\) denotes the estimated \(\alpha^{\prime}\) value required to attain the desired precision and recall scores respectively at a 50\% confidence threshold.
The resultant \(\alpha\) value can be used to subsequently fine-tune the \(\alpha = 1\) model to optimize for the intended metric, where further confidence thresholding can be performed to account for small discrepancies between the predicted \(\alpha\) equations and the observed metrics.

\subsection{Zero-shot woody segmentation}\label{section_woodyDifficulties}
For the zero-shot woody segmentation task, a common difficulty experienced across all the post image generation techniques were large areas of continuous woody vegetation not being detected. All generation techniques excluding the zero image method excelled at detecting isolated groups or rows of trees adjacent to non-woody scenery as seen in Fig.~\ref{fig_woodyExample} in section~\ref{section_woodyDetectionExperiment}. However as the surrounding contextual information denoting the appearance of non-woody vegetation diminishes in the prior image, the predictions of woody vegetation within these areas also diminish. 

This phenomenon is theorized to be caused due to the training dataset for woody clearing not containing examples of large continuous extents of clearing. 
These large continuous woody extents do not provide sharp distinct features such as edges which are commonly extracted by convolutional kernels to help discriminate between classes. 
Instead the homogeneous textures present in thick woody vegetation are similar to the homogeneity of wet grass paddocks and large bodies of water which can subsequently dry up between years, leading to the misleading appearance of woody clearing within confined image context windows. 
This visual similarity towards the negative class and the lack of large continuous examples of the positive class leads the model to resist producing predictions in regions of non-distinct features.
By introducing isolated woody vegetation patches into the post image as reference, the model is grounded by providing many local comparisons as to what woody and non-woody features look like within the current context window.  

The success of the greatest performing zero-shot woody segmentation approach, clearing scene + trees, is dependent on the user’s ability to construct artificial features to extract latent information from the model. 
Constructing an artificial post-image to extract woody segmentation information from a woody change detection model is feasible due to woody clearing leaving distinct land scar features and that discrete examples of woody vegetation for grounding the model can be easily extracted from existing imagery. 
However for more abstract change detection tasks such as post disaster assessment \cite{brunner2010earthquake, zheng2021building, qing2022operational} or general land cover / land use change \cite{chen2024objformer, seyam2023identifying, zhu2022land, das2022land}, creating such post-images to extract latent information may not be practical due to the difficulty associated with quantifying features such as infrastructure damage or a specific land use type into a discrete image form.
In order for such zero-shot post-image augmentation techniques to not be limited by the user’s ability to construct an artificial image capable of extracting the desired latent information, the non-handcrafted approach using activation maximization was proposed at the cost of a reduction in zero-shot performance.

\subsection{Zero-shot woody regrowth detection}
The proposed zero-shot approach enabled the detection of woody vegetation regrowth by utilizing a model trained exclusively on land clearing data, circumventing the need to acquire regrowth labels.

The benefits of the zero-shot approach for global forest monitoring are significant. Given the availability of land clearing datasets and Sentinel-2 imagery, the proposed methodology offers a solution replicable globally. Regions with an established land clearing monitoring program can expand their monitoring capabilities from deforestation to reforestation in a cost-effective and scalable approach without the need to collect new regrowth annotations.

Accurate regrowth detection is critical for climate mitigation strategies, as it supports assessments of carbon sequestration, biodiversity recovery, and forest resilience, potentially correcting for underestimates in current carbon accumulation rates \cite{ cook2020mapping}.

\subsection{Model comparison}\label{section_modelComparison}

\begin{figure}
\centering
\includegraphics[width=1.0\columnwidth]{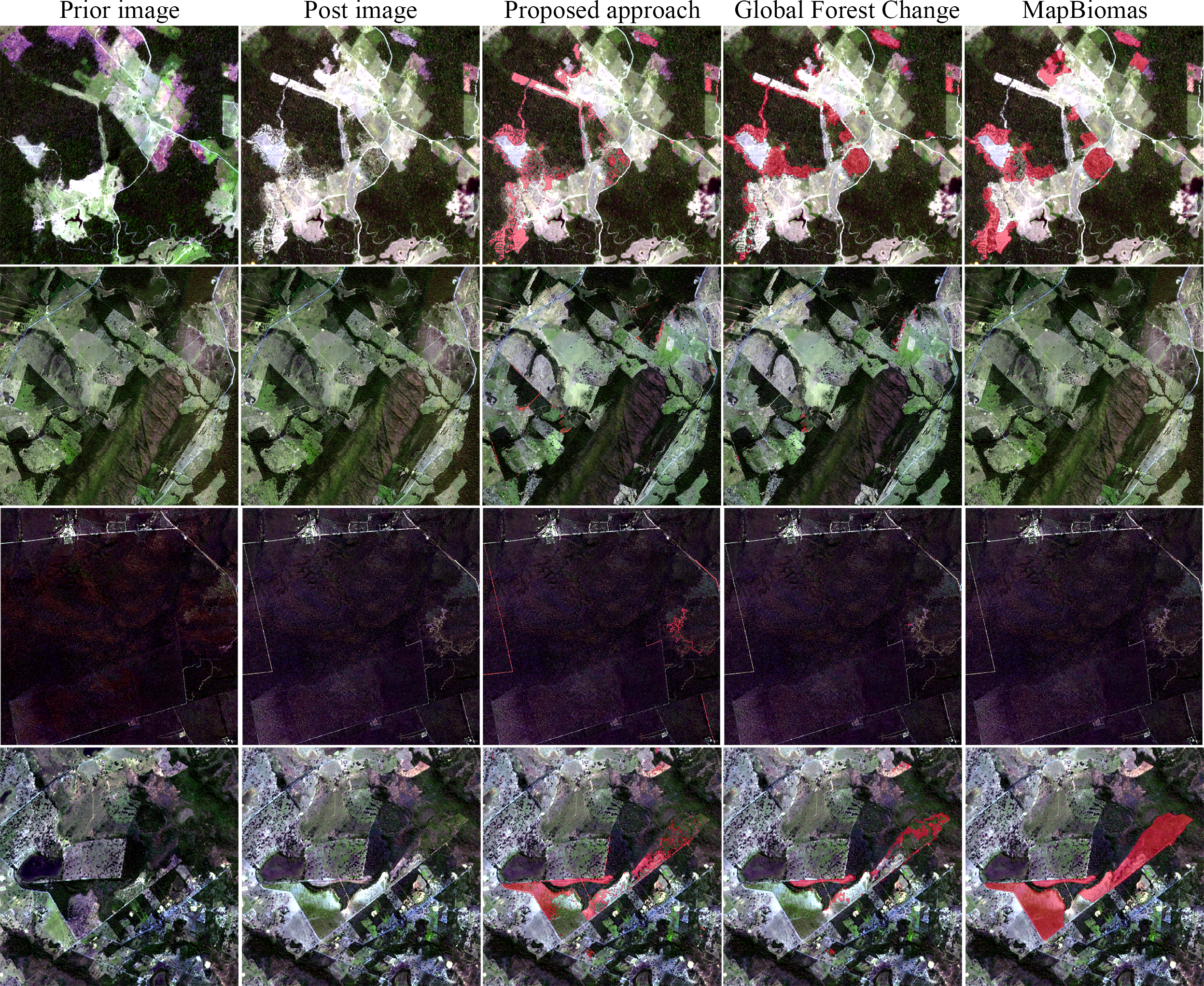}
\caption{Example woody clearing predictions of the proposed approach, Global Forest Change \cite{hansen2013high} and MapBiomas \cite{souza2020reconstructing} overlayed onto the Sentinel-2 post image.}
\label{fig_clearing_comparison}
\end{figure}

Images comparing the proposed approach, Global Forest  Change \cite{hansen2013high} and MapBiomas \cite{souza2020reconstructing} can be seen in Fig.~\ref{fig_clearing_comparison}. The proposed approach excelled at small, thin areas of clearing resultant from roads and paths which often go undetected by the prior methods. Compared to prior works which tend to propose large continuous regions as clearing, the proposed work only flags pixels associated with woody clearing in areas that have undergone clearing operations. It was observed that the proposed model experienced a degradation in performance around areas of cloud cover. Although a cloud mask was used to ignore predictions associated with cloud cover and cloud shadow, the out of distribution features associated with the high and low intensity pixel values from bright clouds and dark shadows resulted in predictions in the immediate neighboring area to be incorrect. The exclusive use of high-quality cloud free imagery during the training stage resulted in the model becoming unrobust to such perturbations. Future work should supplement training datasets with a small sample of images containing cloudy imagery to climatize the model to would-be out of distribution features.

\begin{figure}
\centering
\includegraphics[width=1.0\columnwidth]{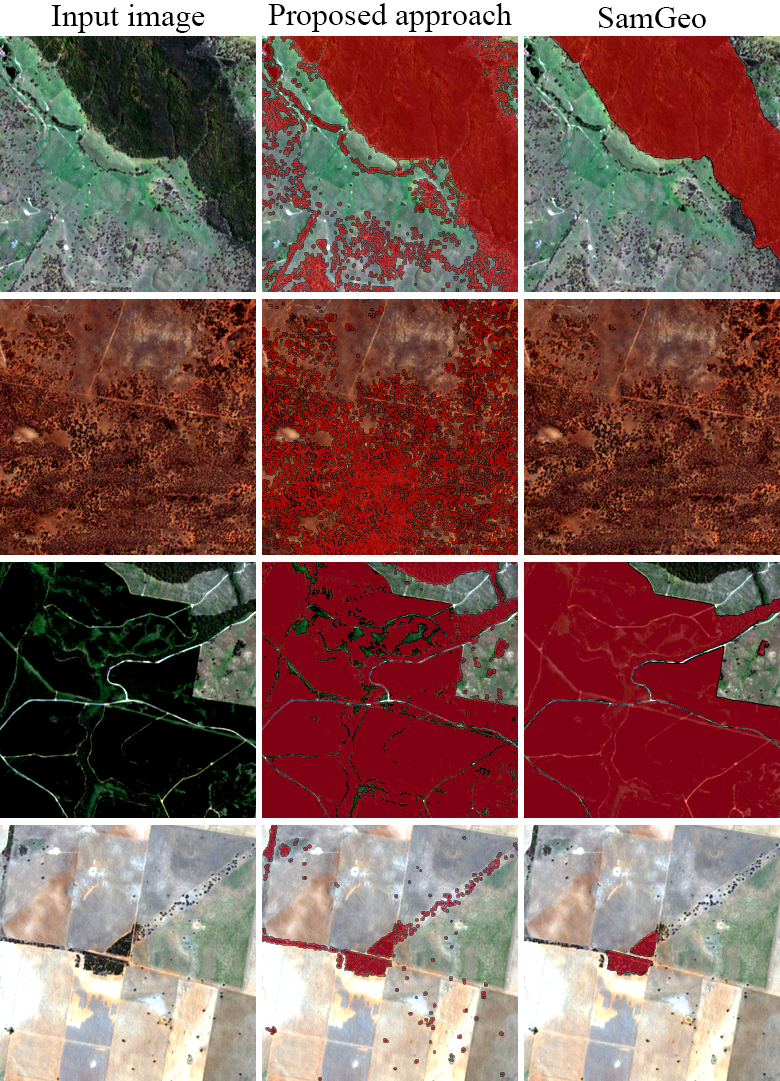}
\caption{Example woody segmentation predictions of the proposed approach and SamGeo \cite{osco2023segment} overlayed on Sentinel-2 imagery.}
\label{fig_woody_comparison}
\end{figure}

Images comparing the predictions of the proposed approach and SamGeo \cite{osco2023segment} in the zero-shot woody segmentation task can be seen in Fig.~\ref{fig_woody_comparison}. The main limitation observed in the predictions generated by the SamGeo \cite{osco2023segment} model was the tendency to favor macro features such as dense forested regions whilst ignoring smaller features such as individual tree crowns. The proposed approach did not suffer from such discrepancies in the generated predictions and was able to correctly segment woody vegetation across all spatial scales.

\begin{figure}
\centering
\includegraphics[width=1.0\columnwidth]{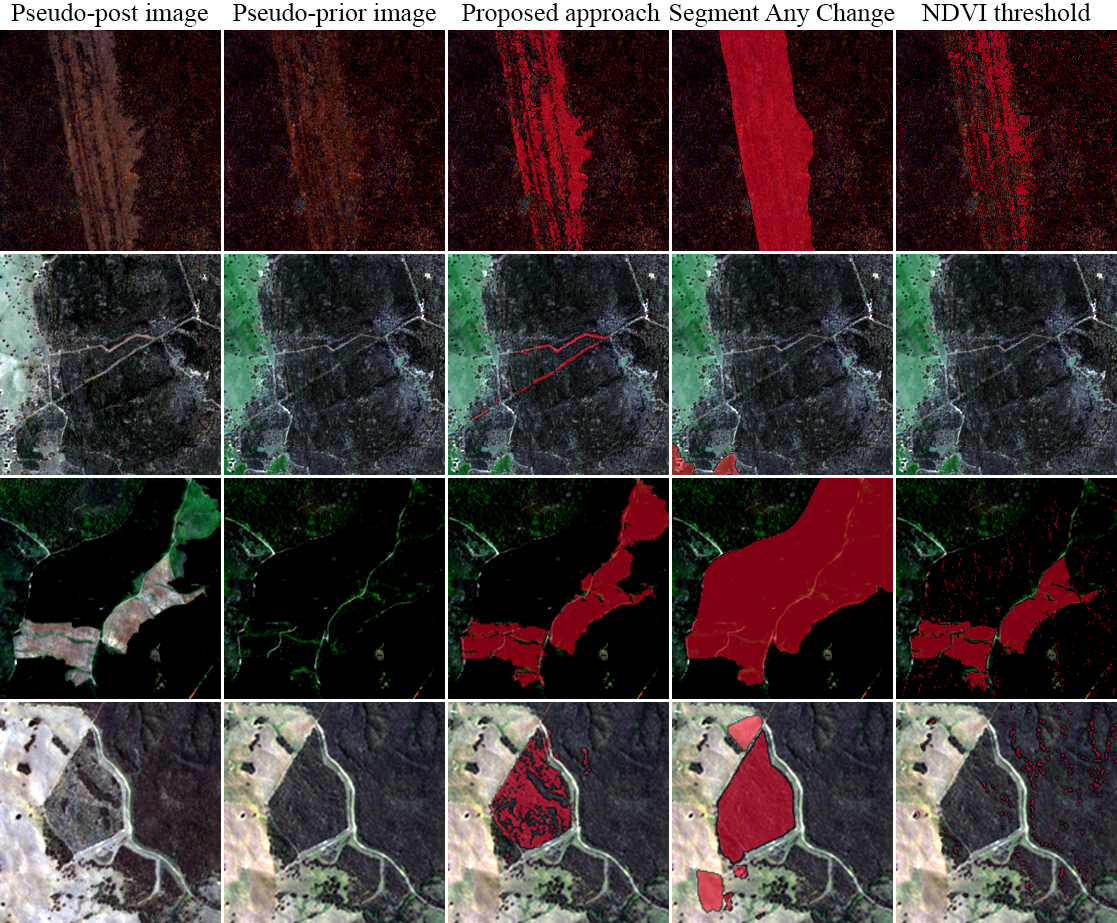}
\caption{Example woody regrowth predictions of the proposed approach, Segment Any Change \cite{zheng2024segment} and NDVI difference thresholding overlayed onto the Sentinel-2 Pseudo-prior image.}
\label{fig_regrowth_comparison}
\end{figure}

Images comparing the predictions of the proposed approach, Segment Any Change \cite{zheng2024segment} and NDVI difference thresholding in the zero-shot woody regrowth task can be seen in Fig.~\ref{fig_regrowth_comparison}. Similar to SamGeo \cite{osco2023segment}, Segment Any Change \cite{zheng2024segment} was observed to favor macro features and over segment large continuous regions. The baseline NDVI difference thresholding method demonstrated spatial consistency issues with salt and pepper like noise in its generated predictions due to the lack of spatial information pooling. The NDVI difference thresholding method required a large discrepancy between NDVI pixel intensities from image pairs containing recently cleared areas and subsequently high density regrowth from plantations to operate effectively.

Across the zero-shot woody segmentation and zero-shot woody regrowth tasks, both SamGeo \cite{osco2023segment} and Segment Any Change \cite{zheng2024segment} experienced difficulties segmenting small scale features. 
Both models utilize SAM \cite{kirillov2023segment} to implement zero-shot segmentation which is trained on a large dataset of 11 million images sourced from photographers. Despite SAM’s \cite{kirillov2023segment} impressive performance on traditional datasets, remote sensing imagery presents alternative challenges not contained within ground-level photography, where coarse 10m Sentinel-2 imagery does not provide the sharp distinct features shown within ground-level photography. Furthermore ground-level photography is constrained to traditional RGB image bands whereas remote sensing imagery leverages hyperspectral imagery to compute indices such as NDVI, NDWI and NDBI to aid in vegetation, water and urbanization classification. Higher resolution imagery that can provide greater detail and more complex features to match ground-level photography may aid such models in attaining greater performance.

\subsection{Limitations}
The main limitation of the proposed work is that it is trained within a constrained geographical location, the state of New South Wales, Australia. Although the proposed work was evaluated outside the training study area, further work is required to both train and evaluate the proposed work in a greater global context to detect woody clearing, regrowth and woody segmentation.

Further limitations of the woody clearing architecture include the exclusion of clearing events deriving from cultivated vegetation such as plantations as discussed in section \ref{section_precisionDiscussion}. Further work is required to filter such data points to attain a more accurate understanding of the model’s precision statistic and to further promote the model’s ability to detect woody clearing events.

An additional limitation of the proposed work is the relatively small sample size evaluation dataset used to assess the zero-shot performance. Compared to the evaluation dataset for woody clearing which used an entire years’ worth of imagery, the evaluation dataset for zero-shot regrowth and woody segmentation amounted to less than 1\% of the total samples used for woody clearing. Further work is required to establish a greater evaluation dataset for woody regrowth and segmentation to more rigorously assess the reliability and performance of the proposed work to operate in zero-shot conditions. 

A further limitation of the proposed work was found within the zero-shot woody segmentation task where the model experienced difficulties predicting regions of large continuous forested regions due to the lack of contextual information. To alleviate this issue, future work should employ self-attention layers as seen in vision transformer architectures \cite{li2022densely, fang2023changer, chen2021remote} to more efficiently pool spatial information, alongside greater training image sizes so that broadscale features can be captured.

The final limitation of the proposed work is the expected performance of zero-shot models compared to supervised models of comparable model complexity and dataset size. The main benefit of zero-shot transfer is the absence of needing to attain high quality input output annotation pairs to train a separate model, which may be costly or practically infeasible to attain. It is recommended that zero-shot transfer should not be used as a main source of generating remote sensing products, with the exception for cases where it is infeasible to attain sufficient training labels, but as a method of providing additional supplementary products at no additional cost.


\section{Conclusion}
In this work we propose a model to detect woody change from input bitemporal Sentinel-2 imagery. We introduce the loss scaling coefficient \(\alpha\) that modifies the binary cross-entropy and dice loss to favor the optimization of precision or recall to better align with end user specifications. Compared to traditional confidence thresholding, the \(\alpha\) conditioned models were able to attain a 1.85\(\times\) higher precision or 1.12\(\times\) higher recall on the woody change dataset.

The proposed work demonstrates input image generation techniques that allow the woody change detection model to zero-shot transfer to woody segmentation tasks with no additional training through visual prompting. 
The most optimal handcrafted post image generation techniques were those that used a mosaic of clearing patches whilst introducing additional artificially pasted trees for contextual grounding.
Automated post image generation techniques using activation maximization performed comparatively to the best performing handcrafted post image generation technique.
Both handcrafted and automated post image generation techniques outperformed the \cite{fisher2016large} Fisher et al. model on the \cite{fisher2016large} Fisher et al. dataset, reducing the error by 18.2\% and 3.8\% respectively.

The proposed woody change detection model was further zero-shot transferred to the woody regrowth detection task where it attained an F1 score of 0.845. The proposed work provides a foundation for future woody regrowth detection work, where due to the rarity and ambiguity of woody regrowth, acquiring datasets to formulate and validate models are difficult.  

\section{Acknowledgments }
Labels and support provided by Vegetation Monitoring \& Reporting Team, Remote Sensing and Regulatory Mapping. Computational power was provided by the Science Data and Compute facility. Both are part of the Science and Insights Division of the New South Wales Department of Climate Change, Energy, the Environment and Water.

\appendices
\section{Woody clearing \(\alpha\)-confidence tables}\label{appendix_alphaConf}
The precision and recall scores for each model conditioned on \(\alpha\) across confidence thresholds ranging from 5\% and 95\% can be seen in tables \ref{table_alphaConfPrecision} and \ref{table_alphaConfRecall} respectively.

\newcommand*\rot{\rotatebox{90}}
\definecolor{TableGray}{gray}{0.9}

\begin{table*}
\centering
\caption{Woody clearing precision}
\begin{tabular}{crccccccccccccccc}\label{table_alphaConfPrecision}
& & \multicolumn{15}{c}{\large \(\alpha\)} \\
& & \(\frac{1}{10}\) & \(\frac{1}{8}\) & \(\frac{1}{6}\) & \(\frac{1}{4}\) & \(\frac{1}{3}\) & \(\frac{1}{2}\) & \(\frac{3}{4}\) & 1.00 & \(\frac{4}{3}\) & 2.0 & 3.0 & 4.0 & 6.0 & 8.0 & 10.0 \\
\midrule
\multirow{19}{*}{\rot{\large Confidence}}
& 5\% & 0.706 & 0.673 & 0.640 & 0.562 & 0.503 & 0.423 & 0.331 & 0.286 & 0.240 & 0.182 & 0.135 & 0.107 & 0.077 & 0.063 & 0.052 \\
& \cellcolor{TableGray}10\% & \cellcolor{TableGray}0.719 & \cellcolor{TableGray}0.687 & \cellcolor{TableGray}0.652 & \cellcolor{TableGray}0.570 & \cellcolor{TableGray}0.510 & \cellcolor{TableGray}0.429 & \cellcolor{TableGray}0.337 & \cellcolor{TableGray}0.291 & \cellcolor{TableGray}0.245 & \cellcolor{TableGray}0.186 & \cellcolor{TableGray}0.139 & \cellcolor{TableGray}0.110 & \cellcolor{TableGray}0.080 & \cellcolor{TableGray}0.066 & \cellcolor{TableGray}0.055 \\
& 15\% & 0.728 & 0.698 & 0.662 & 0.580 & 0.520 & 0.439 & 0.345 & 0.299 & 0.252 & 0.192 & 0.144 & 0.114 & 0.084 & 0.069 & 0.058 \\
& \cellcolor{TableGray}20\% & \cellcolor{TableGray}0.737 & \cellcolor{TableGray}0.708 & \cellcolor{TableGray}0.671 & \cellcolor{TableGray}0.588 & \cellcolor{TableGray}0.528 & \cellcolor{TableGray}0.447 & \cellcolor{TableGray}0.353 & \cellcolor{TableGray}0.306 & \cellcolor{TableGray}0.258 & \cellcolor{TableGray}0.198 & \cellcolor{TableGray}0.148 & \cellcolor{TableGray}0.118 & \cellcolor{TableGray}0.087 & \cellcolor{TableGray}0.072 & \cellcolor{TableGray}0.061 \\
& 25\% & 0.745 & 0.718 & 0.682 & 0.599 & 0.540 & 0.458 & 0.364 & 0.316 & 0.268 & 0.206 & 0.155 & 0.124 & 0.092 & 0.076 & 0.065 \\
& \cellcolor{TableGray}30\% & \cellcolor{TableGray}0.758 & \cellcolor{TableGray}0.734 & \cellcolor{TableGray}0.699 & \cellcolor{TableGray}0.616 & \cellcolor{TableGray}0.558 & \cellcolor{TableGray}0.475 & \cellcolor{TableGray}0.378 & \cellcolor{TableGray}0.330 & \cellcolor{TableGray}0.280 & \cellcolor{TableGray}0.215 & \cellcolor{TableGray}0.163 & \cellcolor{TableGray}0.131 & \cellcolor{TableGray}0.097 & \cellcolor{TableGray}0.080 & \cellcolor{TableGray}0.068 \\
& 35\% & 0.766 & 0.742 & 0.708 & 0.624 & 0.566 & 0.483 & 0.385 & 0.336 & 0.286 & 0.221 & 0.168 & 0.135 & 0.100 & 0.083 & 0.071 \\
& \cellcolor{TableGray}40\% & \cellcolor{TableGray}0.772 & \cellcolor{TableGray}0.749 & \cellcolor{TableGray}0.715 & \cellcolor{TableGray}0.631 & \cellcolor{TableGray}0.573 & \cellcolor{TableGray}0.490 & \cellcolor{TableGray}0.392 & \cellcolor{TableGray}0.342 & \cellcolor{TableGray}0.291 & \cellcolor{TableGray}0.225 & \cellcolor{TableGray}0.172 & \cellcolor{TableGray}0.139 & \cellcolor{TableGray}0.103 & \cellcolor{TableGray}0.086 & \cellcolor{TableGray}0.074 \\
& 45\% & 0.778 & 0.755 & 0.722 & 0.639 & 0.582 & 0.499 & 0.400 & 0.350 & 0.299 & 0.232 & 0.177 & 0.143 & 0.107 & 0.089 & 0.077 \\
& \cellcolor{TableGray}50\% & \cellcolor{TableGray}0.785 & \cellcolor{TableGray}0.764 & \cellcolor{TableGray}0.731 & \cellcolor{TableGray}0.649 & \cellcolor{TableGray}0.592 & \cellcolor{TableGray}0.508 & \cellcolor{TableGray}0.409 & \cellcolor{TableGray}0.359 & \cellcolor{TableGray}0.307 & \cellcolor{TableGray}0.239 & \cellcolor{TableGray}0.183 & \cellcolor{TableGray}0.148 & \cellcolor{TableGray}0.111 & \cellcolor{TableGray}0.093 & \cellcolor{TableGray}0.080 \\
& 55\% & 0.791 & 0.771 & 0.739 & 0.657 & 0.601 & 0.519 & 0.419 & 0.369 & 0.316 & 0.247 & 0.189 & 0.154 & 0.116 & 0.097 & 0.084 \\
& \cellcolor{TableGray}60\% & \cellcolor{TableGray}0.799 & \cellcolor{TableGray}0.780 & \cellcolor{TableGray}0.748 & \cellcolor{TableGray}0.667 & \cellcolor{TableGray}0.612 & \cellcolor{TableGray}0.529 & \cellcolor{TableGray}0.429 & \cellcolor{TableGray}0.378 & \cellcolor{TableGray}0.325 & \cellcolor{TableGray}0.255 & \cellcolor{TableGray}0.196 & \cellcolor{TableGray}0.160 & \cellcolor{TableGray}0.121 & \cellcolor{TableGray}0.101 & \cellcolor{TableGray}0.087 \\
& 65\% & 0.806 & 0.787 & 0.755 & 0.674 & 0.620 & 0.537 & 0.436 & 0.385 & 0.331 & 0.260 & 0.200 & 0.164 & 0.124 & 0.104 & 0.090 \\
& \cellcolor{TableGray}70\% & \cellcolor{TableGray}0.810 & \cellcolor{TableGray}0.792 & \cellcolor{TableGray}0.760 & \cellcolor{TableGray}0.679 & \cellcolor{TableGray}0.625 & \cellcolor{TableGray}0.542 & \cellcolor{TableGray}0.441 & \cellcolor{TableGray}0.390 & \cellcolor{TableGray}0.336 & \cellcolor{TableGray}0.265 & \cellcolor{TableGray}0.204 & \cellcolor{TableGray}0.168 & \cellcolor{TableGray}0.127 & \cellcolor{TableGray}0.107 & \cellcolor{TableGray}0.093 \\
& 75\% & 0.817 & 0.799 & 0.768 & 0.688 & 0.635 & 0.552 & 0.451 & 0.399 & 0.345 & 0.272 & 0.211 & 0.173 & 0.132 & 0.111 & 0.097 \\
& \cellcolor{TableGray}80\% & \cellcolor{TableGray}0.822 & \cellcolor{TableGray}0.805 & \cellcolor{TableGray}0.774 & \cellcolor{TableGray}0.696 & \cellcolor{TableGray}0.643 & \cellcolor{TableGray}0.561 & \cellcolor{TableGray}0.460 & \cellcolor{TableGray}0.408 & \cellcolor{TableGray}0.353 & \cellcolor{TableGray}0.279 & \cellcolor{TableGray}0.217 & \cellcolor{TableGray}0.179 & \cellcolor{TableGray}0.136 & \cellcolor{TableGray}0.115 & \cellcolor{TableGray}0.100 \\
& 85\% & 0.827 & 0.810 & 0.780 & 0.702 & 0.649 & 0.567 & 0.466 & 0.414 & 0.359 & 0.285 & 0.221 & 0.183 & 0.140 & 0.118 & 0.104 \\
& \cellcolor{TableGray}90\% & \cellcolor{TableGray}0.831 & \cellcolor{TableGray}0.815 & \cellcolor{TableGray}0.785 & \cellcolor{TableGray}0.706 & \cellcolor{TableGray}0.654 & \cellcolor{TableGray}0.573 & \cellcolor{TableGray}0.471 & \cellcolor{TableGray}0.419 & \cellcolor{TableGray}0.364 & \cellcolor{TableGray}0.289 & \cellcolor{TableGray}0.225 & \cellcolor{TableGray}0.187 & \cellcolor{TableGray}0.143 & \cellcolor{TableGray}0.122 & \cellcolor{TableGray}0.108 \\
& 95\% & 0.835 & 0.822 & 0.792 & 0.711 & 0.659 & 0.577 & 0.475 & 0.423 & 0.368 & 0.293 & 0.229 & 0.191 & 0.147 & 0.128 & 0.118 \\
\bottomrule
\end{tabular}
\end{table*}

\begin{table*}
\centering
\caption{Woody clearing recall}
\begin{tabular}{crccccccccccccccc}\label{table_alphaConfRecall}
& & \multicolumn{15}{c}{\large \(\alpha\)} \\
& & \(\frac{1}{10}\) & \(\frac{1}{8}\) & \(\frac{1}{6}\) & \(\frac{1}{4}\) & \(\frac{1}{3}\) & \(\frac{1}{2}\) & \(\frac{3}{4}\) & 1.00 & \(\frac{4}{3}\) & 2.0 & 3.0 & 4.0 & 6.0 & 8.0 & 10.0 \\
\midrule
\multirow{19}{*}{\rot{\large Confidence}}
& 5\% & 0.421 & 0.450 & 0.502 & 0.596 & 0.653 & 0.720 & 0.785 & 0.816 & 0.842 & 0.876 & 0.904 & 0.920 & 0.942 & 0.949 & 0.955 \\
& \cellcolor{TableGray}10\% & \cellcolor{TableGray}0.409 & \cellcolor{TableGray}0.438 & \cellcolor{TableGray}0.491 & \cellcolor{TableGray}0.590 & \cellcolor{TableGray}0.648 & \cellcolor{TableGray}0.716 & \cellcolor{TableGray}0.782 & \cellcolor{TableGray}0.813 & \cellcolor{TableGray}0.839 & \cellcolor{TableGray}0.874 & \cellcolor{TableGray}0.902 & \cellcolor{TableGray}0.918 & \cellcolor{TableGray}0.941 & \cellcolor{TableGray}0.947 & \cellcolor{TableGray}0.953 \\
& 15\% & 0.402 & 0.430 & 0.483 & 0.584 & 0.643 & 0.712 & 0.778 & 0.809 & 0.836 & 0.871 & 0.900 & 0.916 & 0.939 & 0.945 & 0.952 \\
& \cellcolor{TableGray}20\% & \cellcolor{TableGray}0.396 & \cellcolor{TableGray}0.423 & \cellcolor{TableGray}0.476 & \cellcolor{TableGray}0.578 & \cellcolor{TableGray}0.638 & \cellcolor{TableGray}0.707 & \cellcolor{TableGray}0.774 & \cellcolor{TableGray}0.806 & \cellcolor{TableGray}0.833 & \cellcolor{TableGray}0.868 & \cellcolor{TableGray}0.897 & \cellcolor{TableGray}0.914 & \cellcolor{TableGray}0.937 & \cellcolor{TableGray}0.944 & \cellcolor{TableGray}0.950 \\
& 25\% & 0.390 & 0.416 & 0.469 & 0.572 & 0.631 & 0.701 & 0.769 & 0.801 & 0.829 & 0.864 & 0.894 & 0.911 & 0.935 & 0.941 & 0.948 \\
& \cellcolor{TableGray}30\% & \cellcolor{TableGray}0.383 & \cellcolor{TableGray}0.408 & \cellcolor{TableGray}0.461 & \cellcolor{TableGray}0.563 & \cellcolor{TableGray}0.623 & \cellcolor{TableGray}0.693 & \cellcolor{TableGray}0.762 & \cellcolor{TableGray}0.794 & \cellcolor{TableGray}0.822 & \cellcolor{TableGray}0.859 & \cellcolor{TableGray}0.889 & \cellcolor{TableGray}0.907 & \cellcolor{TableGray}0.931 & \cellcolor{TableGray}0.938 & \cellcolor{TableGray}0.945 \\
& 35\% & 0.377 & 0.402 & 0.455 & 0.558 & 0.619 & 0.689 & 0.759 & 0.791 & 0.820 & 0.856 & 0.887 & 0.905 & 0.929 & 0.936 & 0.943 \\
& \cellcolor{TableGray}40\% & \cellcolor{TableGray}0.373 & \cellcolor{TableGray}0.397 & \cellcolor{TableGray}0.450 & \cellcolor{TableGray}0.553 & \cellcolor{TableGray}0.614 & \cellcolor{TableGray}0.685 & \cellcolor{TableGray}0.755 & \cellcolor{TableGray}0.788 & \cellcolor{TableGray}0.816 & \cellcolor{TableGray}0.853 & \cellcolor{TableGray}0.884 & \cellcolor{TableGray}0.903 & \cellcolor{TableGray}0.927 & \cellcolor{TableGray}0.934 & \cellcolor{TableGray}0.941 \\
& 45\% & 0.367 & 0.390 & 0.443 & 0.546 & 0.607 & 0.678 & 0.749 & 0.783 & 0.811 & 0.849 & 0.881 & 0.900 & 0.925 & 0.932 & 0.939 \\
& \cellcolor{TableGray}50\% & \cellcolor{TableGray}0.361 & \cellcolor{TableGray}0.384 & \cellcolor{TableGray}0.436 & \cellcolor{TableGray}0.539 & \cellcolor{TableGray}0.601 & \cellcolor{TableGray}0.672 & \cellcolor{TableGray}0.744 & \cellcolor{TableGray}0.778 & \cellcolor{TableGray}0.807 & \cellcolor{TableGray}0.846 & \cellcolor{TableGray}0.877 & \cellcolor{TableGray}0.897 & \cellcolor{TableGray}0.922 & \cellcolor{TableGray}0.930 & \cellcolor{TableGray}0.937 \\
& 55\% & 0.355 & 0.378 & 0.430 & 0.532 & 0.593 & 0.665 & 0.738 & 0.772 & 0.802 & 0.841 & 0.873 & 0.893 & 0.919 & 0.927 & 0.935 \\
& \cellcolor{TableGray}60\% & \cellcolor{TableGray}0.349 & \cellcolor{TableGray}0.372 & \cellcolor{TableGray}0.423 & \cellcolor{TableGray}0.526 & \cellcolor{TableGray}0.587 & \cellcolor{TableGray}0.659 & \cellcolor{TableGray}0.732 & \cellcolor{TableGray}0.768 & \cellcolor{TableGray}0.798 & \cellcolor{TableGray}0.838 & \cellcolor{TableGray}0.869 & \cellcolor{TableGray}0.890 & \cellcolor{TableGray}0.917 & \cellcolor{TableGray}0.925 & \cellcolor{TableGray}0.933 \\
& 65\% & 0.344 & 0.367 & 0.418 & 0.521 & 0.583 & 0.656 & 0.729 & 0.765 & 0.795 & 0.835 & 0.867 & 0.888 & 0.915 & 0.923 & 0.932 \\
& \cellcolor{TableGray}70\% & \cellcolor{TableGray}0.340 & \cellcolor{TableGray}0.361 & \cellcolor{TableGray}0.413 & \cellcolor{TableGray}0.516 & \cellcolor{TableGray}0.578 & \cellcolor{TableGray}0.651 & \cellcolor{TableGray}0.726 & \cellcolor{TableGray}0.761 & \cellcolor{TableGray}0.792 & \cellcolor{TableGray}0.832 & \cellcolor{TableGray}0.865 & \cellcolor{TableGray}0.885 & \cellcolor{TableGray}0.913 & \cellcolor{TableGray}0.921 & \cellcolor{TableGray}0.930 \\
& 75\% & 0.334 & 0.355 & 0.406 & 0.510 & 0.572 & 0.645 & 0.720 & 0.756 & 0.787 & 0.828 & 0.860 & 0.881 & 0.909 & 0.918 & 0.927 \\
& \cellcolor{TableGray}80\% & \cellcolor{TableGray}0.328 & \cellcolor{TableGray}0.349 & \cellcolor{TableGray}0.400 & \cellcolor{TableGray}0.503 & \cellcolor{TableGray}0.565 & \cellcolor{TableGray}0.639 & \cellcolor{TableGray}0.715 & \cellcolor{TableGray}0.751 & \cellcolor{TableGray}0.783 & \cellcolor{TableGray}0.823 & \cellcolor{TableGray}0.856 & \cellcolor{TableGray}0.877 & \cellcolor{TableGray}0.906 & \cellcolor{TableGray}0.916 & \cellcolor{TableGray}0.925 \\
& 85\% & 0.323 & 0.342 & 0.393 & 0.498 & 0.560 & 0.635 & 0.711 & 0.747 & 0.779 & 0.820 & 0.853 & 0.875 & 0.903 & 0.914 & 0.923 \\
& \cellcolor{TableGray}90\% & \cellcolor{TableGray}0.318 & \cellcolor{TableGray}0.335 & \cellcolor{TableGray}0.386 & \cellcolor{TableGray}0.493 & \cellcolor{TableGray}0.556 & \cellcolor{TableGray}0.631 & \cellcolor{TableGray}0.708 & \cellcolor{TableGray}0.744 & \cellcolor{TableGray}0.777 & \cellcolor{TableGray}0.818 & \cellcolor{TableGray}0.851 & \cellcolor{TableGray}0.872 & \cellcolor{TableGray}0.901 & \cellcolor{TableGray}0.911 & \cellcolor{TableGray}0.920 \\
& 95\% & 0.311 & 0.325 & 0.377 & 0.488 & 0.551 & 0.627 & 0.705 & 0.742 & 0.774 & 0.816 & 0.849 & 0.870 & 0.899 & 0.907 & 0.915 \\
\bottomrule
\end{tabular}
\end{table*}

\section{Zero-shot regrowth and woody segmentation \(\alpha\) conditioned precision and recall}\label{appendix_precisionRecallEstimation}
\begin{figure}
\centering
\includegraphics[width=1.0\columnwidth]{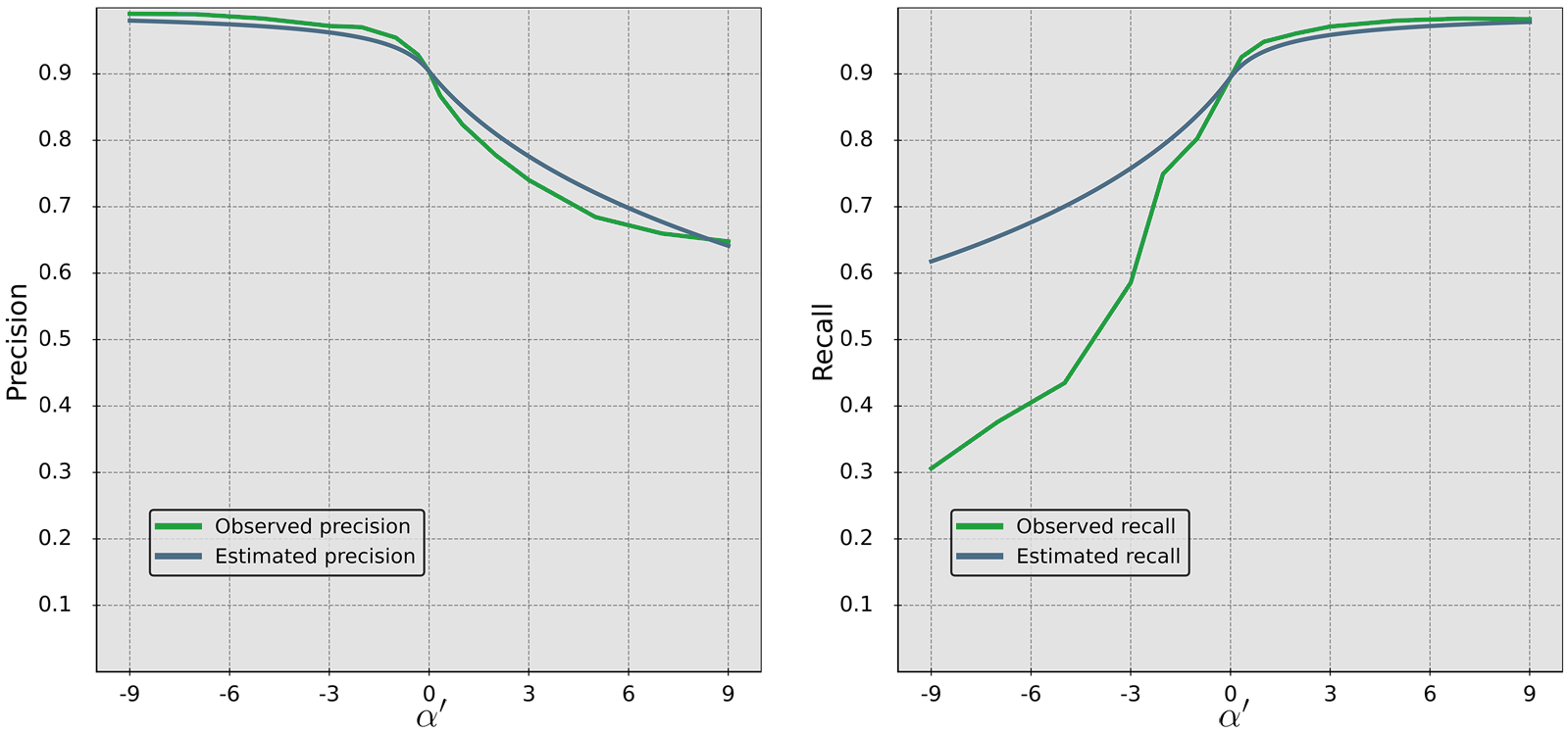}
\caption{Plots comparing the observed precision and recall metrics for the zero-shot woody segmentation task to that of the estimated precision and recall. Observed values were taken using the clearing scene + trees post image generation technique}
\label{fig_alphaPrecisionRecallRegression_woody}
\end{figure}

\begin{figure}
\centering
\includegraphics[width=1.0\columnwidth]{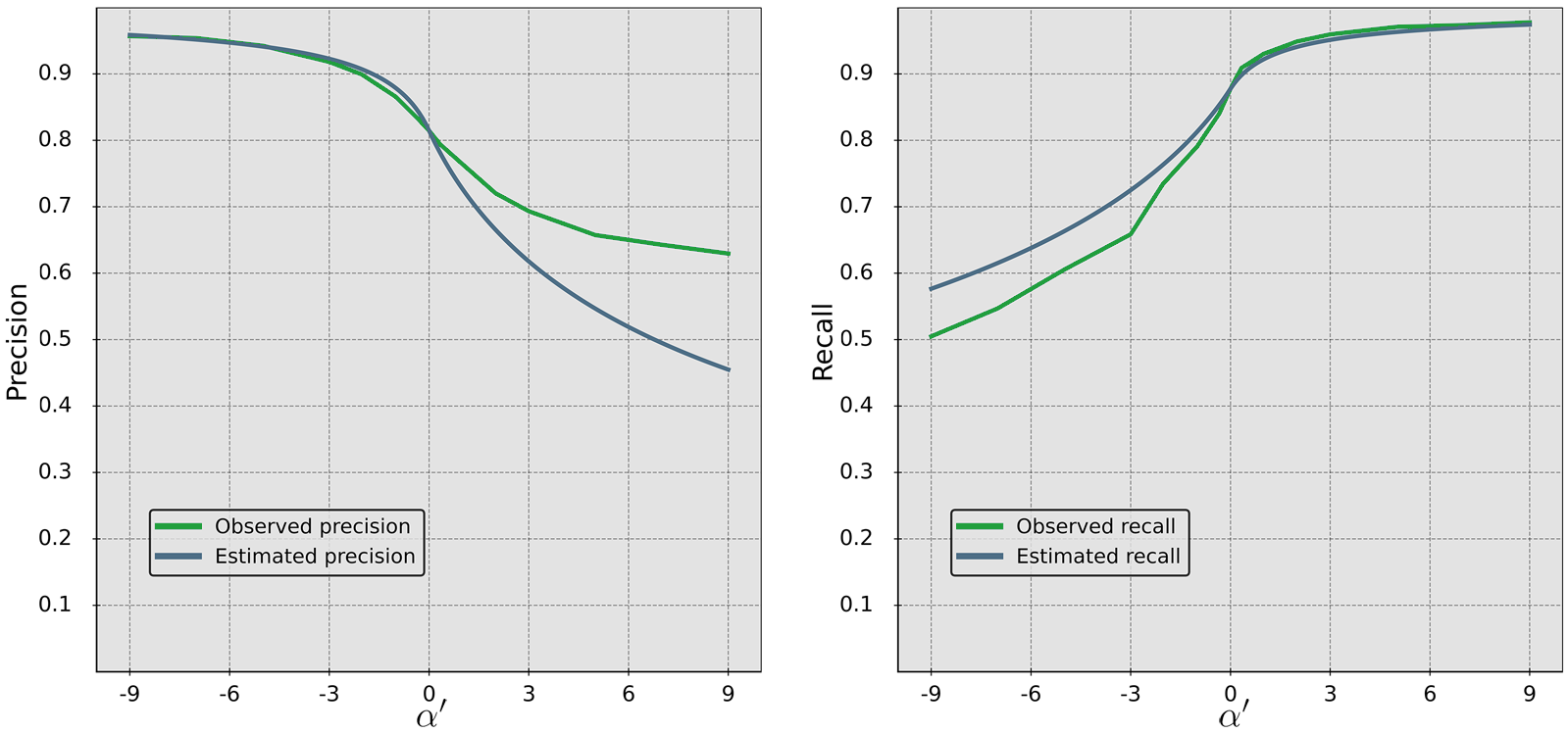}
\caption{Plots comparing the observed precision and recall metrics for the zero-shot woody regrowth detection task to that of the estimated precision and recall. \newline}
\label{fig_alphaPrecisionRecallRegression_regrowth}
\end{figure}

Plots comparing the observed precision and recall scores against predicted precision and recall scores for the zero-shot woody segmentation and regrowth detection tasks can be seen in Fig.~\ref{fig_alphaPrecisionRecallRegression_woody} and Fig.~\ref{fig_alphaPrecisionRecallRegression_regrowth} respectively.

The average error between the observed points and the predicted points for precision and recall for the zero-shot woody segmentation task was 0.017 and 0.079.
The average error between the observed points and the predicted points for precision and recall for the zero-shot woody regrowth task was 0.043 and 0.025.
The average error across all 3 tasks in estimating precision and recall was 0.028 and 0.040 respectively.

The majority of the observed errors for the predicted recall originate from lower \(\alpha^\prime\) values from the woody segmentation task. Observing Fig.~\ref{fig_alphaPrecisionRecallRegression_woody} it can be seen that at \(\alpha^\prime\) values less than -2, a sharp decrease in the recall performance can be observed.
This is due to \(\alpha^\prime\) values less than 0 favouring precision over recall, making the model less susceptible to the input generation techniques used to extract latent woody vegetation information from the model. This sudden drop in recall performance due to the breakdown of the input image generation technique is further exacerbated by the experienced difficulties generating predictions in large areas of continuous woody vegetation as discussed in section \ref{section_woodyDifficulties}. As these large continuous woody vegetation areas constitute a large proportion of the positive class, a slight decrease in their performance results in a large decrease in overall recall.  

\section{Scaled loss comparison}\label{appendix_scaledLossComparison}
The Tversky loss is an alternative loss function that scales the false positives and false negatives to favor precision or recall and is defined as:
\begin{equation}\label{Equation_Tverksy}
L_{Tverky} = 1 - \frac{\sum\limits_{i=1}^{n} Y_{i} \hat{Y}_i}{\sum\limits_{i=1}^{n} Y_{i} \hat{Y}_i + \varphi \sum\limits_{i=1}^{n} (1-Y_i) \hat{Y}_i + \gamma \sum\limits_{i=1}^{n} Y_i (1-\hat{Y}_i)},
\end{equation}
where larger \(\varphi\) values promote a reduction in false positives leading to higher precision scores whilst higher \(\gamma\) values promote a reduction in false negatives leading to higher recall scores. The use of two parameters \(\varphi\) and \(\gamma\) results in a greater hyperparameter search space to find an optimal configuration, where altering either \(\varphi\) or \(\gamma\) impacts both precision and recall due to precision and recall being a tradeoff of one another.    
Compared to the proposed loss function defined in equations~\ref{Equation_alphaDice} \& \ref{Equation_alphaBCE}, a single parameter \(\alpha\) is used to control the preference of recall or precision, leading to a clear signal as to what direction the hyperparameter \(\alpha\) should be moved in to attain the desired performance.

\begin{figure}
\centering
\includegraphics[width=1.0\columnwidth]{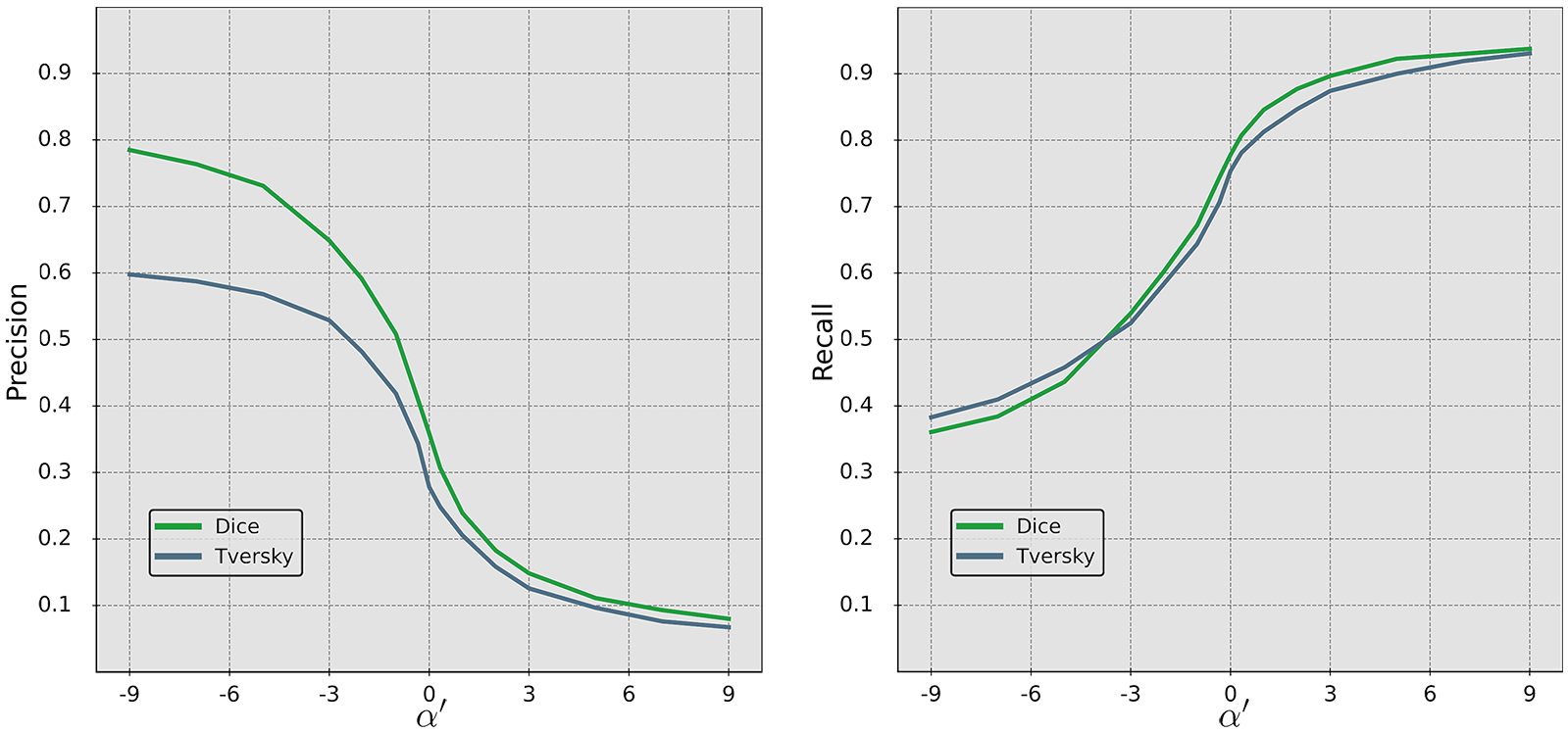}
\caption{Plots comparing the proposed loss and the Tversky loss for the observed precision and recall metrics in the woody clearing detection task.}
\label{fig_alphaPrecisionRecall_Tversky}
\end{figure}

To observe the difference in behavior between the proposed loss function and the Tversky loss, additional models were trained following the process outlined in section~\ref{section_lossScalingTraining} where the dice loss was replaced with the Tversky loss. Due to the difficulty of searching a 2-dimensional state space compared to a 1-dimensional state space, the Tversky loss parameters \(\varphi\) and \(\gamma\) were constrained such that such that \(\varphi = \frac{1}{\alpha}\) and \(\gamma = \alpha\) using the \(\alpha\) values used in section~\ref{section_lossScalingTraining}.
The results of the loss comparison experiment can be seen in Fig.~\ref{fig_alphaPrecisionRecall_Tversky}.

Observing Fig.~\ref{fig_alphaPrecisionRecall_Tversky} it can be seen that the proposed loss function attains marginally greater recall scores whilst experiences significantly greater precision scores for lower \(\alpha^\prime\) values.

\newpage

\ifCLASSOPTIONcaptionsoff
  \newpage
\fi

\bibliographystyle{IEEEtran}
\bibliography{bibtex/bib/IEEEexample}

\end{document}